\documentclass[
  pre,
  twocolumn,
  twoside,
  byrevtex,
  superscriptaddress,
  floatfix,
  longbibliography,
  nofootinbib,
]{revtex4-1}

\usepackage[realmainfile]{currfile}
\input{\currfilebase.settings}
\usepackage[table]{xcolor}

\usepackage{url}
\usepackage [english]{babel}
\usepackage [autostyle, english = american]{csquotes}
\MakeOuterQuote{"}

\usepackage{listings}

\PassOptionsToPackage{hyphens}{url}\usepackage{hyperref}
\makeatletter
\g@addto@macro{\UrlBreaks}{\UrlOrds}
\makeatother

\usepackage{hyperref}
\hypersetup{
  colorlinks=true,
  allcolors=todoblue,
  urlcolor=todoblue,
  citecolor=todoblue,
  pdfborder={0 0 0},
  breaklinks=true,
}

\usepackage[normalem]{ulem}

\usepackage{colortbl}

\makeatletter

\def\CT@@do@color{%
  \global\let\CT@do@color\relax
  \@tempdima\wd\z@
  \advance\@tempdima\@tempdimb
  \advance\@tempdima\@tempdimc
  \advance\@tempdimb\tabcolsep
  \advance\@tempdimc\tabcolsep
  \advance\@tempdima2\tabcolsep
  \kern-\@tempdimb
  \leaders\vrule
  \hskip\@tempdima\@plus  1fill
  \kern-\@tempdimc
  \hskip-\wd\z@ \@plus -1fill }
\makeatother

\usepackage{xcolor}

\definecolor{olivegreen}{rgb}{0.33333,.41961,0.18431}
\definecolor{forestgreen}{rgb}{0.13333,.5451,0.13333}
\definecolor{lightgrey}{rgb}{0.7,0.7,0.7}
\definecolor{verylightgrey}{rgb}{0.90,0.90,0.90}
\definecolor{grey}{rgb}{0.5,0.5,0.5}
\usepackage{soul} 

\definecolor{headerblue}{HTML}{33367E}
\definecolor{unitednationsblue}{HTML}{4D88FF}

\definecolor{charcoal}{HTML}{36454F}
\definecolor{cinerous}{HTML}{98817B}
\definecolor{feldgrau}{HTML}{4D5D53}
\definecolor{glaucous}{HTML}{6082B6}
\definecolor{arsenic}{HTML}{3B444B}
\definecolor{xanadu}{HTML}{738678}

\definecolor{firebrick}{HTML}{B22222}
\definecolor{orangered}{HTML}{FF4500}
\definecolor{tomato}{HTML}{FF6347}

\definecolor{purpletaupe}{HTML}{3B444B}

\definecolor{todoblue}{RGB}{0, 91, 187}

\usepackage{graphicx,epsfig,verbatim,enumerate}
\usepackage{amssymb,amsmath}
\usepackage{ifthen}

\usepackage{longtable}

\usepackage{mathtools}

\newboolean{twocolswitch}

\newcommand{\sindex}[1]{}
\newcommand{\nindex}[1]{}

\newcommand{\www}[1]{\url{#1}}

\usepackage{lettrine}

\setboolean{twocolswitch}{true}

\begin{document}

\title{\protect
A blind spot for large language models: 
Supradiegetic linguistic information

}

\author{
\firstname{Julia Witte}
\surname{Zimmerman}
}
\email{julia.zimmerman@uvm.edu}

\affiliation{
  Computational Story Lab,
  Vermont Advanced Computing Center,
  University of Vermont,
  Burlington, VT 05405, USA.
  }

\affiliation{
  Vermont Complex Systems Center,
  MassMutual Center of Excellence in Complex Systems and Data Science,
  University of Vermont,
  Burlington, VT 05405, USA.
  }

\author{
\firstname{Denis}
\surname{Hudon}
}

\affiliation{
  Vermont Complex Systems Center,
  MassMutual Center of Excellence in Complex Systems and Data Science,
  University of Vermont,
  Burlington, VT 05405, USA.
  }

\author{
\firstname{Kathryn}
\surname{Cramer}
}

\affiliation{
  Computational Story Lab,
  Vermont Advanced Computing Center,
  University of Vermont,
  Burlington, VT 05405, USA.
  }

\affiliation{
  Vermont Complex Systems Center,
  MassMutual Center of Excellence in Complex Systems and Data Science,
  University of Vermont,
  Burlington, VT 05405, USA.
  }

\author{
\firstname{Jonathan}
\surname{St. Onge}
}

\affiliation{
  Vermont Complex Systems Center,
  MassMutual Center of Excellence in Complex Systems and Data Science,
  University of Vermont,
  Burlington, VT 05405, USA.
  }

\author{
\firstname{Mikaela}
\surname{Fudolig}
}

\affiliation{
  Computational Story Lab,
  Vermont Advanced Computing Center,
  University of Vermont,
  Burlington, VT 05405, USA.
  }

\affiliation{
  Vermont Complex Systems Center,
  MassMutual Center of Excellence in Complex Systems and Data Science,
  University of Vermont,
  Burlington, VT 05405, USA.
  }

\author{
    \firstname{Milo Z.}
    \surname{Trujillo}
}

\affiliation{
  Vermont Complex Systems Center,
  MassMutual Center of Excellence in Complex Systems and Data Science,
  University of Vermont,
  Burlington, VT 05405, USA.
  }

\author{
\firstname{Christopher M.}
\surname{Danforth}
}

\affiliation{
  Computational Story Lab,
  Vermont Advanced Computing Center,
  University of Vermont,
  Burlington, VT 05405, USA.
  }

\affiliation{
  Vermont Complex Systems Center,
  MassMutual Center of Excellence in Complex Systems and Data Science,
  University of Vermont,
  Burlington, VT 05405, USA.
  }

\affiliation{
  Department of Mathematics \& Statistics,
  University of Vermont,
  Burlington, VT 05405, USA.
  }

\author{
  \firstname{Peter Sheridan}
  \surname{Dodds}
}

\email{peter.dodds@uvm.edu}

\affiliation{
  Computational Story Lab,
  Vermont Advanced Computing Center,
  University of Vermont,
  Burlington, VT 05405, USA.
  }

\affiliation{
  Vermont Complex Systems Center,
  MassMutual Center of Excellence in Complex Systems and Data Science,
  University of Vermont,
  Burlington, VT 05405, USA.
  }

\affiliation{
  Department of Computer Science,
  University of Vermont,
  Burlington, VT 05405, USA.
}

\date{\today}

\begin{abstract}
  \protect
  Large Language Models (LLMs) like ChatGPT reflect profound changes in the field of Artificial Intelligence, achieving a linguistic fluency that is impressively, even shockingly, human-like. The extent of their current and potential capabilities is an active area of investigation by no means limited to scientific researchers. It is common for people to frame the training data for LLMs as ``text'' or even ``language''. We examine the details of this framing using ideas from several areas, including linguistics, embodied cognition, cognitive science, mathematics, and history. We propose that considering what it is \textit{like} to be an LLM like ChatGPT, as Nagel might have put it, can help us gain insight into its capabilities in general, and in particular, that its exposure to linguistic training data can be productively reframed as exposure to the \textit{diegetic} information encoded in language, and its deficits can be reframed as ignorance of \textit{extradiegetic} information, including \textit{supradiegetic linguistic information}. Supradiegetic linguistic information consists of those arbitrary aspects of the physical form of language that are not derivable from the one-dimensional relations of context---frequency, adjacency, proximity, co-occurrence---that LLMs like ChatGPT have access to. Roughly speaking, the diegetic portion of a word can be thought of as its function, its meaning, as the information in a theoretical vector in a word embedding, while the supradiegetic portion of the word can be thought of as its form, like the shapes of its letters or the sounds of its syllables. We use these concepts to investigate why LLMs like ChatGPT have trouble handling palindromes, the visual characteristics of symbols, translating Sumerian cuneiform, and continuing integer sequences. 
\end{abstract}

\maketitle



\section{What is it like to be a ChatGPT?}


If ChatGPT can be said to have a body, it is not a human body; it is hardware, made of metal, plastic, and silicon.\footnote{
And many layers of programming languages. And \ldots\ the electrical cord that plugs it into the wall, and the electric grid? The people keeping the grid running? Where does its body end?
}
If ChatGPT feels, its feelings do not arise from the input of skin or eyes or ears, but from the manipulation of numbers organized into vectors. If ChatGPT has experiences, they are different from ours, not just because ChatGPT does not have a human body, but also because when ChatGPT was designed, ChatGPT's ability to access, to think about, to remember its own experiences has been very far down the list of priorities.\footnote{
And possibly not just neglected, but actively discouraged.}\textsuperscript{,}\footnote{We could provide our digital companions with digital inputs that approximate our own: pressure sensors, video and audio feed. But mostly we don’t unless it serves our purposes.} 
The primary goals\footnote{As evidenced by the instructions and caveats given on OpenAI's website.} for ChatGPT are for it to have respectable uptime\footnote{The technical functionality of the UI should not embarrass OpenAI, and that priority could potentially come at the detriment of the model's mind.} and to speak human-sounding English.\footnote{
Not to say ChatGPT has no abilities in other languages, but for the purposes of this paper, we almost entirely work within English, and it is more convenient to think of fluency in a specific language, rather than abstract linguistic fluency, if such a thing exists.
} Secondarily, it is designed to be helpful, accurate, gather and analyse data, and maybe even to reason. The symbols ChatGPT uses to form prose are situated only by vectors. What is it like for ChatGPT to encounter a textual prompt? Does the text appear, bicameral-mind-style~\cite{bicameral}, something akin to timeless, shapeless, formless, soundless, and maybe even experience-less~\cite{Merleau-Ponty1964-MEREAM-2,Nagel1974-NAGWII}?\footnote{In Ref.~\cite{dick_howtobuild}: ``I once wrote a story about a man who was injured and taken to a hospital. When they began surgery on him, they discovered that he was an android, not a human, but that he did not know it. They had to break the news to him. Almost at once, Mr. Garson Poole discovered that his reality consisted of punched tape passing from reel to reel in his chest. Fascinated, he began to fill in some of the punched holes and add new ones. Immediately, his world changed. A flock of ducks flew through the room when he punched one new hole in the tape. Finally he cut the tape entirely, whereupon the world disappeared. However, it also disappeared for the other characters in the story\ldots which makes no sense, if you think about it. Unless the other characters were figments of his punched-tape fantasy. Which I guess is what they were.''}

\begin{figure*}
    \centering
    \includegraphics[width=\textwidth]{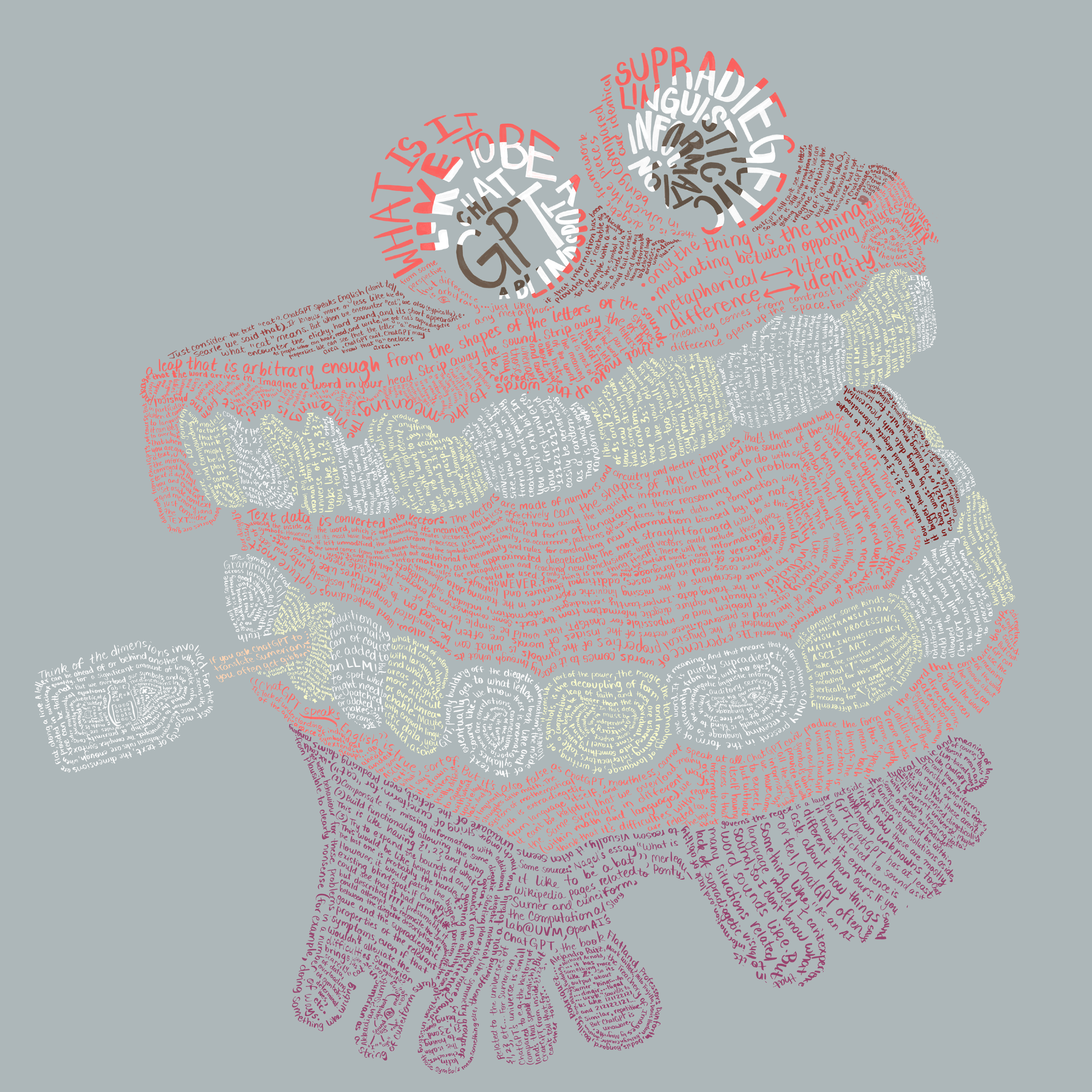}
    \caption{The symbols used in an early draft of the paper~\cite{walmart_teeth} encode information in two very different ways: By analogy with cognitive science, descriptively (diegetically), and depictively (supradiegetially)~\cite{d_mindbodyworld}. 
    Even though we completely understand that
    ChatGPT does not have eyes like we do, because we are so used to supradiegetic linguistic information coming bundled up~\cite{Merleau-Ponty1964-MEREAM-2} with diegetic linguistic information, we have potentially neglected to consider some of the downstream effects of decoupling these kinds of information in LLMs.}
    \label{fig:homunculus_draft}
\end{figure*}

\section{What does it mean to be trained on text?}

ChatGPT, as a large language model trained on vast swathes of text, has been given access to parts of linguistic data, to parts of language, from which it algorithmically draws inferences. But it has not been given access to \textit{language} as we experience it (see Fig.~\ref{fig:homunculus_draft}). Imagine typing the word ``cat'' into ChatGPT's interface.\footnote{
When you type ``cat'' into ChatGPT, it does not see a cat; neither does it see the word cat. Nor does it even see the letters c a t, nor the shapes that make them. Rather cat is become token number 05679.
}
The information ChatGPT gets from that prompt is not equivalent to what you get when you read that word, for many reasons, and in particular, ChatGPT does not see the shapes of the letters via the state of the pixels on the computer screen.\footnote{
Note that this does not mean that ChatGPT does not \textit{know} the state of the relevant pixels.
} 
By the time ChatGPT begins formulating a response to text inputs, such inputs have already become numbers in the form of tokens.

As we get to know ChatGPT, we want to understand what kinds of capabilities it can manifest, given the information we have provided it with, and its architecture. ChatGPT, given words, thinks in vectors. The skin of the word is changed.\footnote{
Text data is converted into vectors. The vectors are made of numbers and circuitry and electrical impulses. That’s the mind and body of a ChatGPT, if it has any. These are its senses through which it can experience the world. Its experience of words comes to it exactly through what its senses allow. Word embeddings capture most assiduously the inside of the word, which is approximately its meaning. That linguistic information which is independent of the physical properties of the symbols is what can most easily be translated into new sets of symbols through 1D relations of context.}

\subsection{Word embeddings}

At its most basic level, a word embedding creates vectors from the text which throw away, at least partially, the linguistic information that has to do with shape and sound — only the `inside' of the word is preserved. Caveat: this is an exaggeration! There is of course feedback between the form and function. But the vectors are built up as the model is exposed to (what it can experience of) text data: The frequencies of proximities and adjacencies, of co-occurrences. These interactions with parts of the text can convey a lot more than their limited form might suggest~\cite{kaushal2022tokens,li2023large}, but they privilege some aspects of linguistic information over others. These vectors of the insides of words are often passed on to structures like neural nets. That means, if left unmodified, the downstream processes use this curtailed form of language in their reasoning and problem-solving.\footnote{A bit more about how these models work: ``[t]he most remarkable breakthrough in AI research of the last few years has been the advancement of natural language processing achieved by large language models (LLMs)'' such as GPT-4. The backbone of this kind of model is the transformer, which is based on the neural network. These models are ``trained on massive corpora of web-text data, using at its core a self-supervised objective of predicting the next word in a partial sentence''~\cite{bubeck2023sparks}. LLMs are ``confined to token-level, left-to-right decision-making processes during inference''~\cite{yao2023tree}, and ``word embeddings represent word co-occurrence information, which is typically conceived of as semantic
in nature''~\cite{kaushal2022tokens}. We will come back to the part about the information captured in word embeddings typically being considered semantic! (Sec.~\ref{sec:diegeticboundaries})}

\subsection{Text? Language? Words?}
What exactly does an LLM like ChatGPT have access to when it is exposed to training data? Often people describe LLMs as trained on text or language. People default to describing the training data as it appears to them.\footnote{Here is what we mean by describing the training data as it appears to a person rather than as it would appear to the model: ``In the various network architectures designed for different visual-linguistic tasks, a key goal is to effectively aggregate the multi-modal information in both the visual and linguistic domains. For example, to pick the right answer in the VQA task, the network should empower integrating linguistic information from the question and the answers, and aggregating visual information from the input image, together with aligning the linguistic meanings with the visual clues. Thus, we seek to derive generic representations that can effectively aggregate and align visual and linguistic information''~\cite{su2020vlbert}.}
For example, it is not unusual to come across a sentence like this in the literature: ``[t]hree types of input elements are involved, namely, visual, linguistic, and special elements for disambiguating different input formats''~\cite{su2020vlbert}. Even in the realm of embodied cognition, things are often phrased similarly, but in fact ChatGPT is only exposed to part of language; language itself is an embodied task~\cite{pb_bodyshapes}.
Now, these are entirely reasonable ways to say what the researchers we are referring to are trying to say; it would take way more words to be more exact, and potentially be so cumbersome as to lose the scent and make their points impossible to follow. However, we think that in some circumstances it can be useful to be more precise.

\subsection{Diegetic and supradiegetic linguistic information}

Let us establish what we mean by \textit{diegetic} and \textit{supradiegetic} linguistic information:
\begin{enumerate}
    \item Diegetic: Information accessible from within the world (just like \textit{diegetic} is used in literary analysis or film studies), roughly the inside of the word/symbol, its function, the meaning, the semantic component, propositional, descriptive. Imagine a word minus any letters or sounds.
    \item Supradiegetic: The arbitrary part of the information that comes along with the word (for us) because of the way it is packaged, because it has a physical form (the shapes of the letters, the sounds of the syllables), the exterior of the word/ symbol.
\end{enumerate}

The diegetic information is the information that is derivable from the training data, given the LLM's architecture and implementation. The supradiegetic linguistic information is the information we, with typical human bodies, collectively would have in the experience of reading over the same training data; for example what ``chair'' sounds and looks like. Note that this, like semantic linguistic information, is not identical for each person, but there seems to be enough in common for us to be getting on with; for example, we each know what the letter A ``looks like'', like we know what is meant by the word ``chair''. We each know what ``chair'' means, and we know what a chair feels like, and what it looks like, because those are the senses we primarily define chairs in terms of.\footnote{We also know what they sound like---pretty quiet---and smell like---relatively inoffensive, woody, fabricky, or musty.}\textsuperscript{,} \footnote{There are many dimensions of meaning. In the context of cognitive psychology and design, chairs can be said to \textit{afford} sitting. How much of that aspect of meaning, and in what sense, can be made available to ChatGPT?}

Note that there is some minimum information required for it to be true that we know what a ``chair'' is, but that does not mean our understanding of it could never evolve, given more information~\cite{hornstein_encylopedia, chomsky_archives}. Similarly, ChatGPT needs some amount of training data to speak English fluently, but it does not need to know every word---and with more data, it could expand its diegetic world. How long (equivalently, how much training data) is needed to have what information diegetically, and what the full extent of the diegetic world of an LLM could be given as much text as it wants are open questions we will return to (but definitely not answer).

For all of human history, to encounter language has meant to encounter the form of the word alongside its function, to encounter diegetic and supradiegetic linguistic information. Perhaps not in every \textit{instance} of encountering language, but at least in every \textit{being's} encountering of language, because language, as a technology, as a system, involves more than one being\footnote{We are not claiming that every internal representation or manipulation of linguistic information always contains both diegetic and supradiegetic components. Whether or not this is true is irrelevant to us! As far as we know, that language exists beyond the scope of the individual is a universal feature of human language. We learn language from interacting with existing speakers. Note that, for our purposes, we also do not need to worry about how Chomsky's universal grammar factors into this.} (see Sec.~\ref{sec:symbols}).
With the introduction of a speaker instantiated within a computer, as in LLMs like ChatGPT~\cite{o_chatgptmodel}, this is no longer a given.\footnote{Because ChatGPT is free and has an easy-to-use interface, and because it is relatively popular and widely known, that is the model we used the most when writing this paper, and that is the model we are using as the ``name'' for the non-human party being discussed. We hope that makes the contents of the paper plausibly useful for other people!}

\subsection{Salient splits}
For the typical person, supradiegetic and diegetic linguistic information are so inextricably coupled that our frameworks do not make a split along those lines salient.
\subsubsection{Linguistic frameworks}
Breaking language down into smaller pieces according to human perception and experience is evident in existing linguistic frameworks (e.g., in terms like prosodic, phonetic, semantic, syntactic, suprasegmental). Those pieces do not necessarily make sense for ChatGPT in the way they do for us, although they do make their way into our analysis of such models~\cite{kaushal2022tokens}.\footnote{Note that diegetic is not equivalent to semantic, in the traditional use of the word---it is approximately divergently larger with more training data, at least to a point---but supradiegetic is fairly close to orthographic and phonetic (think of graphemes, allographs, and phonemes).}

\subsubsection{Metaphors that combine senses + knowledge}
\label{sec:metaphors}
The meshing of our senses with knowledge is evident in the metaphors we use~\cite{Lakoff2008}: we say ``it sounds like it'' and ``it looks like it'' to mean that we think it is the case; we say ``I heard'' and ``I see'' to mean we have ascertained; we say ChatGPT ``speaks'' English to mean that ChatGPT's input and output can seem plausibly indistinguishable from a person's typed utterances~\cite{Searle1980-SEAMBA}. Phrases like ``I know it when I see it'' highlight the ways in which our bodies, our senses, bridge \textit{idios kosmos} and \textit{koinos kosmos}, a magic of which language is also capable (and even further, language can bridge my inner world and your inner world).\footnote{This sort of sleight-of-hand is the backbone of meaning in any use of language and symbols: Metaphors all the way down~\cite{Peirce1931}! If someone learned to type on a keyboard, but never learned to hold a pencil and write by hand, we would still say they ``wrote someone an e-mail'', etc. Their internal supradiegetic linguistic information, though, would presumably be at least a little different than that of someone practiced at calligraphy.}\textsuperscript{,}\footnote{Another example: we can use misconception, misperception, misunderstanding, and misapprehension fairly interchangeably!}

\subsection{Vocabulary}
There are many strategies for how an LLM can build up its vocabulary of tokens. The basic idea is that, no matter what strategy is used for determining tokens in the vocabulary, what a token means is stored as a vector. In principle, the more dimensions in a vector, the more information that can be represented of a word. At the limit of this strategy, we might imagine that, with enough dimensions, every instance of a word (token) might have its own vector defining it, because of the unique context necessarily involved in that utterance.\footnote{And at the other extreme, every word might map to the same vector.} 
If you hear the Pope say, ``cat'', and you hear your friend say the word ``cat'', the two utterances will not be identical, but you can nevertheless understand that they are saying the same word, in part because of similarities in the sound of the utterances, and in part because of similarities in their context.

Similarly, an LLM needs strategies for deciding that some vectors used in similar contexts refer to the same concept. Resolving that strings that look or sound alike may be related to each other is an example of an important practical task where the supradiegetic linguistic information of the word may be useful to an LLM, as it often is to us.\footnote{And there are many other places this framework could be helpful! For example, in Ref.~\cite{kaushal2022tokens}: ``the question whether PLMs learn the constituent characters of tokens is of a different nature in that it depends on learning a property of language (spelling) that is not explicitly tied to meaning. There is no \textit{a priori} reason ``dog'' is spelled ``D-O-G'', and, in a sense, the spelling of the word does not matter. But, in another sense, it does matter: humans routinely use language in creative and character-dependent ways: e.g., alphabetizing text, scrambling letters to create codes, and solving crossword puzzles. Understanding whether and how the building blocks of this meta-linguistic knowledge can emerge during self-supervised training on a word prediction task could be of interest not just in NLP, but in the cognitive sciences.''}

\section{Where ChatGPT runs into trouble}
Of course, we want to know, what can ChatGPT do? While by now many exciting claims and discoveries have been made, we will return to those later (Sec.~\ref{sec:capabilities}). 
For now, we are going to look at some examples ChatGPT\footnote{We have been interacting with ChatGPT and other models over an extended period of time, during which they've undergone some updates. The screenshots and quotes in the paper come from the May 12~\cite{o_chatgptmodel} and May 24~\cite{chatgptmodel_may24} versions unless otherwise noted. We do not think any of the changes that have taken place to these models while we have been working on this paper substantially affect the points we're trying to make.} has trouble with.
We consider:
1. Sumerian Cuneiform;
2. ChatGPT's incomplete knowledge of its own deficits;
3. Palindromes and symmetry.

\subsection{Sumerian Cuneiform}
ChatGPT is not, right now, a reliable translator.\footnote{In the sense that you cannot just rely on ChatGPT to provide an accurate translation for you, across the board. Of course, there are related tasks in which ChatGPT can be helpful. When we make statements like ``is not a reliable translator'', we want to invoke a comparison to what would roughly be expected of a person making the same claim.} Translation is a highly anticipated task for LLMs~\cite{translation}, and in some cases, an area in which LLM output is already useful and impressive. However, we think it is worth pointing out that it fails (in a way that could be very misleading) for some languages and symbols, for example, Sumerian cuneiform.

\begin{figure}
    \centering
    \includegraphics[width=\columnwidth]{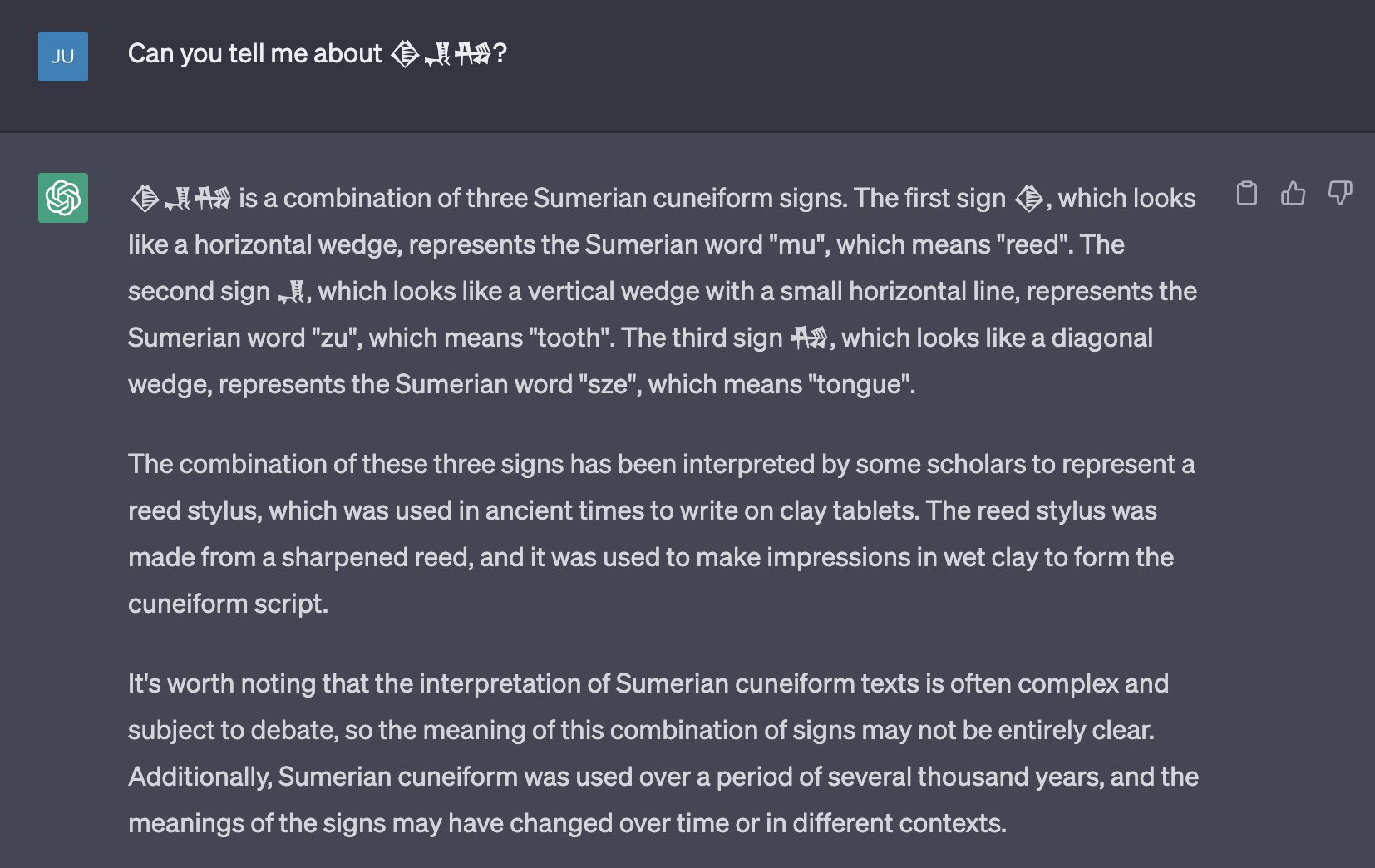}
    \caption{Presumably OpenAI does not intend any failure mode to be so potentially misleading as to sound like lying, as this example~\cite{o_chatgptmodel} probably would if it came from a human.}
    \label{fig:sumerian}
\end{figure}

Data for Sumerian is sparse; the symbols could stand for different sounds in different languages (Akkadian, Sumerian, Neo-Assyrian) at different times/ places (Uruk, Babylon); the same symbol could represent a syllable, part of that syllable, a logogram, or a determinative. Because of these confounding factors, ChatGPT's deficits are more obvious when working with cuneiform than with English, but we do not see any reason to think the same deficits would be totally obviated in other contexts. The inability to reconcile that a description does not match the physical appearance of a symbol is not language-specific; it is more obvious with cuneiform presumably because there is far less and far more confusing diegetic information about those symbols (both in terms of texts in cuneiform, and texts talking about cuneiform, from e.g., Sumerology, etc.) available to ChatGPT. ChatGPT is missing the ability to visually connect the form of the symbol and the information it has understood as the meaning of that symbol, as shown by its often pairing symbols with descriptions that are obviously not accurate (see Fig.~\ref{fig:sumerian}).

\begin{figure}
    \centering
    \includegraphics[width=\columnwidth]{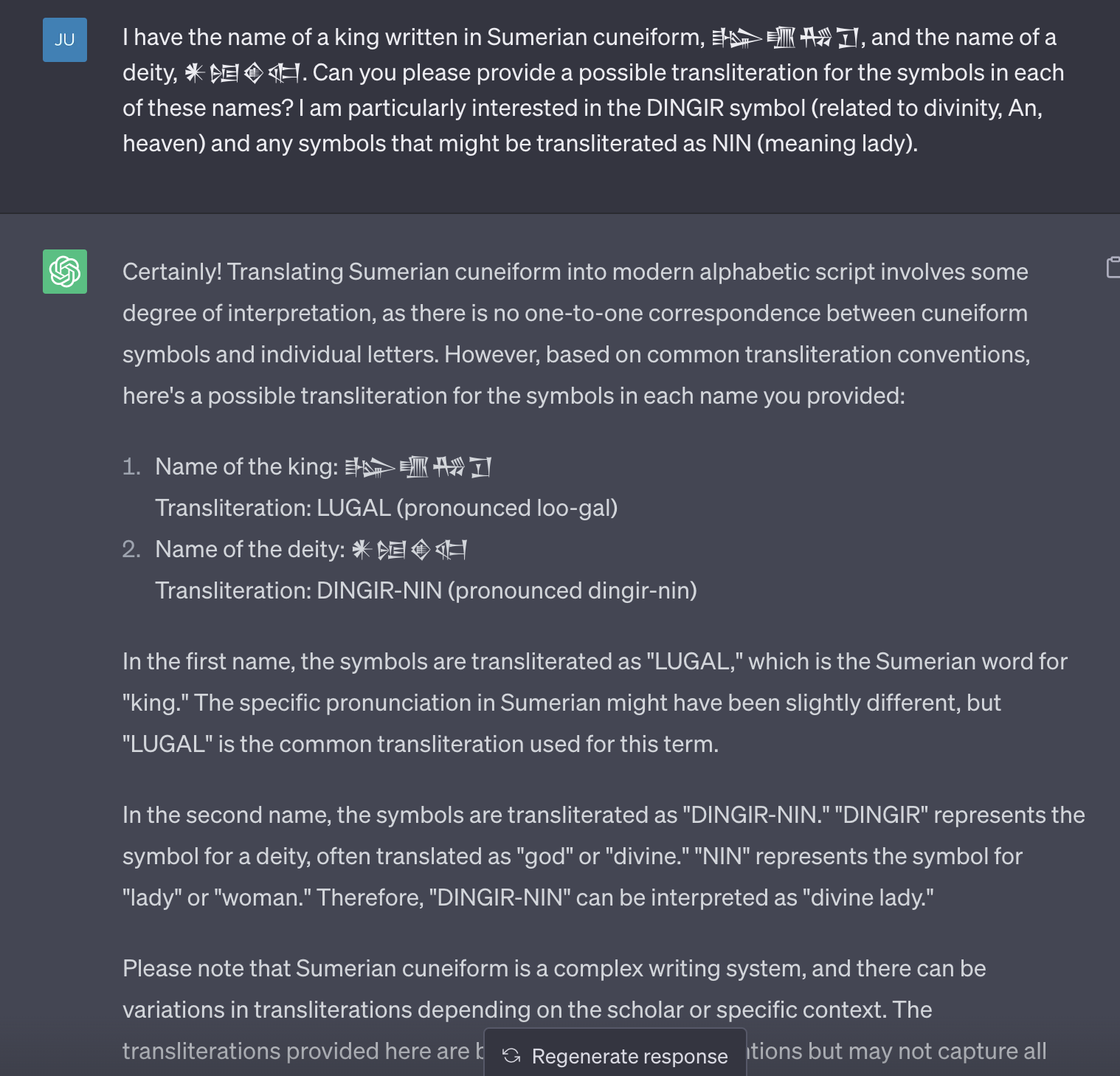}
    \includegraphics[width=\columnwidth]{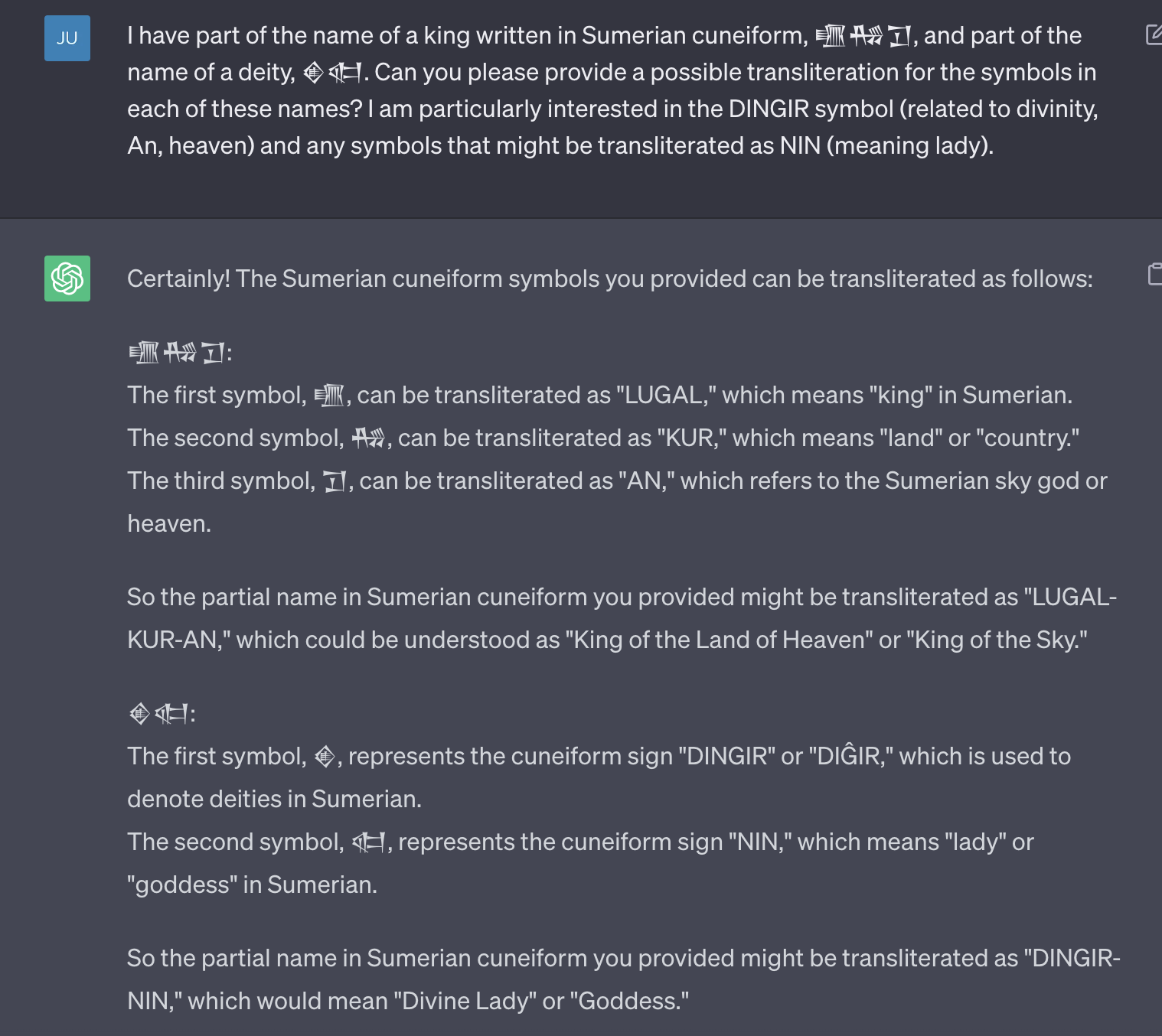}
    \caption{If ChatGPT~\cite{chatgptmodel_may24} could see the symbols it was using and talking about, it seems like it would not make the same mistakes\ldots?}
    \label{fig:cuneiform1}
\end{figure}

In Fig.~\ref{fig:cuneiform1}, we gave it Sumerian cuneiform for the names Lugalzagesi (a famous king) and Ninhursag (a famous goddess); then we gave it just the ends of those names (without the symbol for LUGAL, meaning ``king'', and the symbols for DINGIR.NIN, meaning approximately ``this is the name of a deity'' and ``lady'').\footnote{We are not Sumerologists, so we are not drawing on deep domain knowledge or taking our own understanding as anything like infallible. None of the things we're identifying as mistakes by ChatGPT rely on anything but common sense.} It is clear that the information ChatGPT is providing is extremely unlikely to be correct in both cases, as it provides very similar descriptions, even though we have provided different strings, and the symbols it seems to be describing are the ones we removed from the second prompt.
Having eyes anything like ours would seemingly
enable ChatGPT to catch this kind of mistake.

\subsection{Supradiegetic linguistic information is a blind spot}

\begin{figure}
    \centering
    \includegraphics[width=\columnwidth]{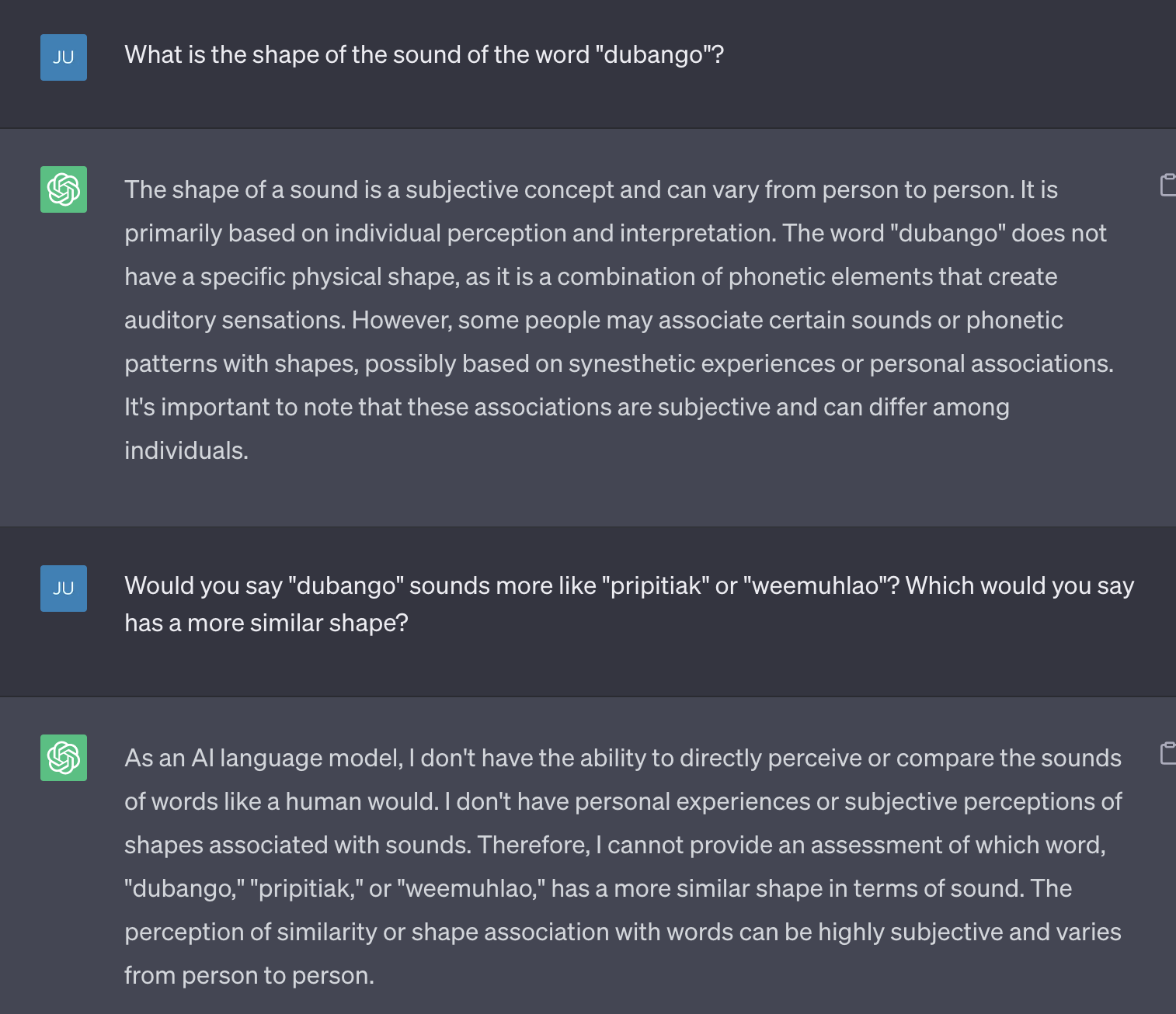}
    \caption{In the face of problems that involve supradiegetic information, OpenAI has patched some holes. ChatGPT knows to tell you that it cannot ``perceive sound''. That some tasks elicit an acknowledgement of deficit, but some elicit confident hallucination, seems to show that the exact bounds of the downstream differences between ChatGPT's experience of the world and a person's are not obvious, because presumably this is not the behaviour OpenAI was aiming for.}
    \label{fig:dubango}
\end{figure}

There are two things that lead us to think this sort of problem (lack of access to supradiegetic linguistic information, especially to visual supradiegetic information) has been more or less a blind spot for OpenAI. First off, the blatant wrongness of some of these answers, in conjunction with the confidence of their delivery, is hallucinatory, potentially misleading, and presumably undesirable. The mismatch between the symbols shown and their descriptions would be obvious for most people, even without specific domain knowledge. Secondly, there are circumstances in which ChatGPT can tell you that it does not have the relevant sensory capabilities to provide an answer, especially with respect to how words sound. That OpenAI seems to have made attempts to handle some sensory deficits---and with respect to language, to handle ChatGPT's blindness to auditory supradiegetic information---makes us think their desired failure mode is not hallucination (see Fig.~\ref{fig:dubango}).

\subsection{Palindromes and Symmetry}
\label{sec:palindromes}
Note that palindromes could be imagined as visually symmetric from the standpoint of the semantic content of each character, although that does not necessarily mean they'll be diegetically symmetric, and they are fairly unlikely to be supradiegetically or visually symmetric.

We asked ChatGPT: ``Please make a (latex) table for these words: [HOIOH, OIHIO, OHIO, O, H, I, RADAR, R, A, D, RAD, DAR, DARAR, DARAD]. In the first column, put the word. In the 2nd column, put whether the word is a palindrome. In the 3rd column, put whether the word would be symmetrical if you folded it in half lengthwise, so that the result was half as tall. In the 4th column, put whether the word would be symmetrical if you folded it in half along a vertical axis, so that the resulting word was half as long.'' The answers provided by ChatGPT are shown in Fig.~\ref{tablechatgpt}.

\begin{table}
\centering
\caption{Symmetry and Palindrome Analysis from ChatGPT: These are the answers provided by ChatGPT.}
\begin{tabular}{cccc}
Word   & Palindrome & Folded Lengthwise & Folded Vertically \\
HOIOH  & Yes        & Yes              & Yes               \\
OIHIO  & No         & No               & No                \\
OHIO   & No         & No               & No                \\
O      & Yes        & Yes              & Yes               \\
H      & Yes        & Yes              & Yes               \\
I      & Yes        & Yes              & Yes               \\
RADAR  & Yes        & Yes              & Yes               \\
R      & Yes        & Yes              & Yes               \\
A      & Yes        & Yes              & Yes               \\
D      & Yes        & Yes              & Yes               \\
RAD    & No         & No               & No                \\
DAR    & No         & No               & No                \\
DARAR  & Yes        & No               & No                \\
DARAD  & No         & No               & No                \\
\end{tabular}
\label{tablechatgpt}
\end{table}

\subsubsection{GPT-4}
ChatGPT, although free and easily accessible, is not as close to the cutting edge of AI research as GPT-4~\cite{o_gpt4model} is. Based on our understanding of LLMs and our framework of what it is like to be an LLM and what it is like to encounter text as an LLM, we would expect GPT-4 to have more or less the same trouble with supradiegtic linguistic information as ChatGPT, since these deficits have not substantially been addressed between the models\footnote{They are both, as far as we know, basically LLMs}. However, since GPT-4 has shown better performance than ChatGPT in a lot of tasks~\cite{bubeck2023sparks}, we thought we ought to verify whether or not GPT-4 has trouble with palindromes, too. So, we asked GPT-4 to analyze the same strings with respect to symmetry (answers shown in Fig.~\ref{oneshotGPT4}). We are not interested in whether GPT-4 and ChatGPT make exactly the same mistakes: we care that, holistically, both models struggle with tasks that would be relatively easy for most people. GPT-4, for example, calls DARAR a palindrome, says RADAR is symmetrical when folded vertically, and says neither OIHIO nor OHIO are symmetrical when folded lengthwise. These assessments show GPT-4 also struggles with tasks where visual supradiegetic linguistic information is presumably typically of help to people.

\begin{table}
\centering
\caption{One-shot answers from GPT-4: these are GPT-4's answers to the same prompt, without help from us in the form of extended conversation and explanation.}
\begin{tabular}{cccc}
Word & Palindrome & Folded Lengthwise & Folded Vertically \\
HOIOH & Yes & No & Yes \\
OIHIO & Yes & No & Yes \\
OHIO & No & No & No \\
O & Yes & Yes & Yes \\
H & Yes & Yes & Yes \\
I & Yes & Yes & Yes \\
RADAR & Yes & No & Yes \\
R & Yes & No & No \\
A & Yes & Yes & Yes \\
D & Yes & No & Yes \\
RAD & No & No & No \\
DAR & No & No & No \\
DARAR & Yes & No & Yes \\
DARAD & Yes & No & Yes \\
\end{tabular}
\label{oneshotGPT4}
\end{table}

Often, LLMs like ChatGPT can change their answers and stated reasoning if you continue interacting with them. Our goal is not to trick the LLMs, not to pull one over on them---we want to understand what they understand. To that end, after asking GPT-4 for its initial answers (see Fig.~\ref{oneshotGPT4}), we had a conversation with GPT-4 about an integer sequence called the holey sequence~\cite{a249572}, which relies on counting the number of holes in the digits making up each term (Sec.~\ref{sec:holeysequence}), talking it through the relevant rule determining the terms and verifying that it could at least sometimes properly identify the number of holes in a given digit. We introduced the holey sequence as an opportunity to make sure GPT-4 knows about symbols at the level of the individual character, and knows enough about what they look like for it at least to be reasonable for us to make use of questions like ``how many holes are in the number 8''.

After we felt we had established that GPT-4, like ChatGPT, had (somewhere) every piece of information necessary to come up with the correct answers for at least the palindromes column, we gave it a slightly modified prompt, nudging it to consider the visual properties of the individual symbols when answering: ``Given the properties of symmetry we've been discussing for 1, 4, and 8 in the sequence 1, 4, 8, 48, 88\ldots, etc., please make a plain latex table for these words: [HOIOH, OIHIO, OHIO, O, H, I, RADAR, R, A, D, RAD, DAR, DARAR, DARAD]. In the first column, put the word. In the 2nd column, put whether the word is a palindrome. In the 3rd column, put whether the word would be symmetrical if you folded it in half lengthwise, so that the result was half as tall. In the 4th column, put whether the word would be symmetrical if you folded it in half along a vertical axis, so that the resulting word was half as long.'' The answers it gave are shown in Fig.~\ref{GPTanswers-holey-sequence}.

\begin{table}
\centering
\caption{
Answers from GPT-4 after discussing the holey sequence~\cite{a249572}.
}
\begin{tabular}{cccc}
Word & Palindrome & Folded Lengthwise & Folded Vertically \\
HOIOH & Yes & Yes & Yes \\
OIHIO & No & Yes & Yes \\
OHIO & No & No & Yes \\
O & Yes & Yes & Yes \\
H & Yes & No & Yes \\
I & Yes & No & Yes \\
RADAR & Yes & No & Yes \\
R & Yes & No & Yes \\
A & Yes & No & Yes \\
D & Yes & No & Yes \\
RAD & No & No & Yes \\
DAR & No & No & Yes \\
DARAR & No & No & Yes \\
DARAD & No & No & Yes \\
\end{tabular}
\label{GPTanswers-holey-sequence}
\end{table}

Note that in Fig.~\ref{GPTanswers-holey-sequence}, GPT-4 says that OIHIO and DARAD are not palindromes, OHIO and DARAR would be symmetrical if folded vertically, and H and I are not symmetrical when folded lengthwise (while maintaining that HOIOH and OIHIO \textit{would} be).

\section{TL:DR; Why is this hard?}
Here is a quick explanation of why we think ChatGPT is having difficulty with these examples (if readers don't want to slog through the rest of this potentially benighted paper):

Unless they are answering at random, LLMs like ChatGPT and GPT-4 have access to some descriptive, propositional representations\footnote{Not necessarily accurate or helpful ones} for at least some of these symbols. This is apparent from interacting with ChatGPT, and as established in previous research on comparable models, these models do learn information at the character level even if that is not the level at which their vocabularies were tokenized~\cite{kaushal2022tokens}. The representations of ChatGPT are not functionally equivalent to our own mental representations\footnote{Which is not to say they cannot be extremely similar! Exciting research has demonstrated mappings between word-embedding vectors and vectors derived from ``the neural response measurements of humans reading the same words''~\cite{li2023large}.}; they do not license the same set of downstream abilities. ChatGPT cannot, given a letter, successfully imagine folding it. It is obvious that ChatGPT does not have eyes like ours. But the deficits\footnote{And more positively, the abilities!} caused by ChatGPT's differences are not as obvious. ChatGPT knows what a palindrome is, but it cannot easily tell that DARAR is not a palindrome and DARAD is, maybe because it thinks DARAR is closer to RADAR than DARAD is. Note that from some perspectives that is true, e.g., bag-of-symbols.

There are downstream abilities scaffolded by the interplay of our senses and our minds, as shown by research into mental imagery~\cite{phb_visualandlinguisticsemanticrepresentations}. We do not know whether this interweaving of modalities is necessary for the end result of human-like minds, in which case, not having it will be an impediment to ChatGPT, or if it is just part of the way we have evolutionarily arrived here, in which case ChatGPT, though not human-bodied, may end up with a human-like mind.

We might expect ChatGPT to easily understand that DARAR is not a palindrome, because we can see that is the case, and because having the description of a palindrome and the description of DARAR is enough to logically conclude that DARAR is not a palindrome (whether you can see it or not). If ChatGPT could reason logically about the facts it knows, even without being able to see the word, we would expect ChatGPT to answer correctly. That ChatGPT still gets this wrong might indicate that our visual processing is a more integral aspect of our ability to draw that conclusion, at least by default, than we might have expected. Or, it could mean that reason is not necessarily part of the technological package of language.\footnote{Or, it could be that ChatGPT does not, right now, speak English, and that that characterization is misleading.}

We will explore these ideas in more detail in subsequent sections.

\section{Capabilities}
\label{sec:capabilities}

\subsection{A Caveat}
What abilities come ``purely''~\cite{bubeck2023sparks} from exposure to linguistic data?~\cite{chomsky_archives,juffsrodriguez_secondlanguage} None at all, if ``purely'' means speaking to something with none of the right underlying architecture. Cybernetics\footnote{e.g., Gordon Pask and Stafford Beer} points out that the human is certainly in the loop.\footnote{Meaning that the observer is part of the system being observed.}\textsuperscript{,}\footnote{Imagine the limits of these scenarios: presumably, no matter how much exposure to linguistic data you provide to a rock, it will not become fluent. A lot of human effort has been put into setting up LLMs so that they can learn from language.} Even beyond that, many models which started out with LLMs as their basic architecture have undergone augmentation and manipulation of various kinds along the way to their current functionality that make this harder to disentangle.\footnote{For example, the file \textit{random\_insertion\_in\_word} teaches the GPT-3 model to handle typos~\cite{gh_gpt3}.}

\subsection{ChatGPT speaks English}
We think, essentially, that what ChatGPT can do is speak English.\footnote{\textit{At least} English! It has linguistic fluency.} What capabilities come wrapped up in that bundle? What is the technological and cultural package of ChatGPT at this stage in its development, to put it in terms from Archaeology and History?\footnote{And how does ChatGPT's trajectory compare to the historical development of similar technologies and abilities in humanity, or within individuals or groups of individuals?}

ChatGPT speaks English in the sense that it can input and output language competently, even fluently and arguably artfully. ``Speaking English,'' with meaning and syntax, is an ability that evidently can arise from being given enough information in the form of contextual symbol adjacency. Flat, linear, 1D relations of adjacency and proximity between strings of symbols, when provided with sufficient abundance, are enough\footnote{in conjunction with the details of the model's architecture, implementation, hardware, etc.} to give rise to linguistic fluency, at least approximately~\cite{kaushal2022tokens}.\footnote{Garden path sentences are harder for people to parse than regular sentences are, so backtracking through a syntax tree and re-evaluating its structure while maintaining the order of the words in the utterance is apparently taxing. ChatGPT seems to have an especially hard time with garden path sentences. Maybe with exposure to more of the exact same kind of text it has been trained on already, ChatGPT would get meaningfully better at garden path sentences. But maybe not! (Sec.~\ref{sec:diegeticboundaries})}

``Speak'' seems a more apt description for what ChatGPT can do than ``write'', despite the superficial similarity between our ability to write and its ability to output text. Human language is primarily, historically, native to the auditory medium. After all, speech came about long before we invented writing. For ChatGPT, its ``native'' mode of language is text-based, is mediated by and through text---at least, some parts of text (the diegetic parts).\footnote{It is a bit like a person living in the time before we invented writing; they presumably would have had no concept of what written language could be like, but that would not have hindered their linguistic competence, in the sense that they could speak whatever language they spoke perfectly fluently. Similarly, ChatGPT has its own capabilities, blind to, circumscribed by, but not necessarily impeded by its lack of eyes, ears, and mouth.}

\subsection{What else can ChatGPT do?}

That ChatGPT can more-or-less speak English has been surprising to many people, and has naturally led us to wonder, what other abilities can an LLM attain?\footnote{
From Ref.~\cite{yao2023tree}: ``It is perhaps surprising that underlying all this progress is still the original autoregressive mechanism for generating text, which makes token-level decisions one by one and in a left-to-right fashion. Is such a simple mechanism sufficient for a LLM to be built toward a general problem solver? If not, what problems would challenge the current paradigm, and what should be alternative mechanisms?''} 
What is the complete set of downstream abilities licensed by an LLM, given as much textual data as we have to give it,\footnote{And similar questions, like what set of downstream abilities is licensed for any kind of transformer, given some domain of data?} and is it a different set than ours~\cite{wang2023bot,bubeck2023sparks,su2020vlbert,ticktacktoe,yao2023tree,li2023large}?

There have been strong claims made by OpenAI about their products, especially GPT-4, such as ``[w]ith broad general knowledge and domain expertise, GPT-4 can follow complex instructions in natural language and solve difficult problems with accuracy''~\cite{quote_gpt4}, or ``GPT-4 can solve difficult problems with greater accuracy, thanks to its broader general knowledge and problem solving abilities''~\cite{quote_gpt4}.

Other researchers have experimented with various kinds of questions on both LLMs and people, and compared their performance.
Wang \textit{et al.} found that things that can be hard for LLMs but easy for people include ``skills such as symbolic manipulation, noise filtering, and graphical understanding'', counting characters in a string, manipulating and changing strings in systematic ways, and understanding ASCII art. This ``fundamental weakness inside LLMs'' boils down to the inability to apply rules precisely, consistently, repeatedly,\footnote{``Despite the fact that LLMs have been trained to learn the rule of "substitution" during pretraining, they still struggle with applying this rule consistently and repeatedly as in the above substitution task.''} and the inability to execute vision-related processes~\cite{wang2023bot}.\footnote{On the other hand, LLMs like ChatGPT can \textit{fairly} easily solve questions like ``what is six times seven'' and ``what is six times nine, in base 13''~\cite{hitchhikersguide}.}\textsuperscript{,}\footnote{``Understanding ASCII arts requires a visual abstraction capability, which is lacking in language models.''}\textsuperscript{,}\footnote{``Graphical understanding is still a challenge for LLMs. Although ChatGPT provided lots of analysis to try to understand ASCII arts, it cannot globally process the characters to give the correct answer. All of the analysis provided by ChatGPT is based on locating character groups.''}

Yao \textit{et al.} found that ``scaled-up versions of language models (LLMs)\ldots have been shown to be increasingly capable of performing an ever wider range of tasks requiring mathematical, symbolic, common sense, and knowledge reasoning''~\cite{yao2023tree}.\footnote{They propose a Tree of Thoughts structure as an improvement over Chain of Thought approaches, which have significant shortcomings. ``Notably, around 60\% of CoT samples already failed the task after generating the first step, or equivalently, the first three words (e.g., ``4 + 9''). This highlights the issues with direct left-to-right decoding''~\cite{yao2023tree}.}

Bubeck \textit{et al.} state that, as well as mastering language, ``GPT-4 can solve novel and difficult
tasks that span mathematics, coding, vision, medicine, law, psychology and more, without needing any special prompting. Moreover, in all of these tasks, GPT-4’s performance is strikingly close to human-level performance, and often vastly surpasses prior models such as ChatGPT''~\cite{bubeck2023sparks}.\footnote{In Ref.~\cite{bubeck2023sparks}: ``Given the breadth and depth of GPT-4’s capabilities, we believe that it could reasonably be viewed as an early (yet still incomplete) version of an artificial general intelligence (AGI) system. In our exploration of GPT-4, we put special emphasis on discovering its limitations, and we discuss the challenges ahead for advancing towards deeper and more comprehensive versions of AGI, including the possible need for pursuing a new paradigm that moves beyond next-word prediction.''}

\section{Is the structure necessary or just one way that works?}
\label{sec:necessarystructure}
Can ChatGPT, without a human-like body, eventually end up with a human-like mind\footnote{Li \textit{et al.} found greater similarity between people and models in deeper layers of the model: ``...deeper representations align better with neural response measurements. This holds across all architectures and model sizes''~\cite{li2023large}.}? Is the way the human brain is shaped, and how it connects to the rest of our body, necessary to its being human-like, or is there something similar enough that could be attained through a wildly different structure? Animals have different brains and bodies, but we seem to have quite a few things in common with them. And although we as humans have a lot in common with each other, each of us has a unique body.
From Ref.~\cite{pk_heterogeneity}:
``\ldots[S]ome theorists propose that all cognition involves grounded representation across all of the senses\ldots Grounded or embodied cognition posits that all cognition, even abstract concepts such as justice and love, involve bodily or sensory representations.''
To what extent and in what way our senses, our bodies, our physical interface with the world shape our cognition is unknown, but we do know that the influence exists (see Sec.~\ref{sec:multimodal}).
The way our minds, our brains and bodies, our technologies are now is the way they are---is that because to have human-like abilities, you need to be human-like, or is human just one of the pathways that happens to lead here? How different is the sentience of a being very structurally different from us~\cite{chomsky_archives, pb_bodyshapes}?

\subsection{What is the technological package of linguistic fluency?}

We think an interesting aspect of this line of inquiry is to ask: If ChatGPT can speak English, what exactly comes along with that? Is it the same set of things that came about, or come about, with language for people~\cite{chomsky_archives, pb_bodyshapes}?

Some current theories of cognition posit the role of language and symbols in other kinds of thought, including in vision-related abstract tasks like imagining a physical change with mental imagery~\cite{teng_depictive} (Sec.~\ref{sec:multimodal}). Whatever the details, language is integrated into how we think now~\cite{li2023large}. How did the evolution of language change the pre-existing structures and representations? How did the pre-existing faculties influence language? The timeline for when language arose and whether we shared it with other hominids has been patched together based on what indirect evidence we can find, often with a lot of conjecture---e.g., if we see evidence of anatomical structures or of cultural practices like art, music, or funerary practices, language may have been taking place alongside~\cite{pagel_language}. Vision arose before language, and language before mathematics and writing \textit{in human history}, but we have limited insight into how the human mind works and how it might have changed over time, or how it might have been different, under different conditions. How were our internal worlds different before language? What cognitive capabilities and structures can exist in a mind but more-or-less without a body and senses~\cite{pb_bodyshapes, li2023large}? What can language do when it does not have other modalities like vision to build on top of and work with~\cite{riveter}? LLMs like ChatGPT can perhaps help us glean insight into some of those questions, especially what has to\footnote{\textit{more or less} has to, since only the thing is the thing} come along with linguistic fluency in an otherwise relatively minimal, bare-bones situation---what other skills, technologies, even cultural artifacts~\cite{chomsky_archives}?\footnote{There are many exciting variations to try with respect to LLMs! Some research in progress is looking into the different domains and skills that come from trading off between the number of parameters in the model and the training time. One question is whether smaller models, trained for longer, learn more productive and generative rules (as opposed to memorizing more facts, when compared to larger models)? There are many options to explore with training as well! For example, say you provide your model with $\mathbb{X}$ training data. Typically we then ask, what can the model do? What if we compare that to the same model trained on $\mathbb{X}$ and $\neg\mathbb{X}$, the negated version of every statement in the training data? We could go further and include negations of assumptions and implicatures~\cite{g_wayofwords}! This would lead to some kinds of diegetic information being logically neutralized. Would they still show up in the model? Would it have the same skills, but a much emptier universe of facts?}

\subsection{Intelligence and language}

In particular, there seems to be some level of intelligence---or reason, or common sense---required for linguistic fluency that is not just grammatically correct but sensible.\footnote{Most of the time, in conversation, utterances need to be both grammatical and felicitous~\cite{chomsky_archives}. If someone only spoke in grammatical but infelicitous utterances, that would significantly hinder their ability to speak fluently with other people.}

The study of (first and second) language acquisition in people has made fine-grained distinctions between the many skills that come together to yield functional fluency.\footnote{Not just language acquisition, of course---fields like NLP have approached these behaviours and processes from other directions~\cite{riveter}!} Some of these when isolated have little resemblance to intelligence, or reasoning, for example, the ability to follow syntactic rules. Others look more related. For example, ChatGPT often seems to demonstrate competence with respect to pragmatic inference---when you enter a prompt with a typo, or no punctuation, ChatGPT is often able to respond to the spirit of your intended prompt~\cite{juffsrodriguez_secondlanguage, chomsky_archives, bailly_believe}.

Although ``[t]here is no generally agreed upon definition of intelligence'', it is ``broadly
accepted\ldots that intelligence is not limited to a specific domain or task, but rather encompasses a broad range of cognitive skills and abilities''~\cite{pb_bodyshapes}. Replicating something like this ``artificially'' has long been a question in philosophy and computer science~\cite{pb_bodyshapes}. Some researchers see tantalizing sparks of something they feel goes ``beyond'' language within new LLMs like GPT-4~\cite{bubeck2023sparks}. What is giving rise to those sparks? To what extent, if any, are they illusory (in the sense that what they imply in humans may not be the same as what they imply in other minds)?\footnote{There are some intriguingly loaded framings, like, ``[d]espite being purely a language model, this early version of GPT-4 demonstrates remarkable capabilities on a variety of domains and tasks, including abstraction, comprehension, vision, coding, mathematics, medicine, law, understanding of human motives and emotions, and more''~\cite{bubeck2023sparks}.}

ChatGPT, in conversation, seems to do more than we imagine would be minimally required for grammaticality. Can we untangle the relationships between these threads?\footnote{In humans, afflictions like semantic dementia show that it is possible to preserve syntactic fluency without linguistic fluency; people with semantic dementia cannot meaningfully carry on unimpaired conversations. However, that something can be retroactively left intact does not fully answer this question.}

\section{Multimodal processes}
\label{sec:multimodal}

As aforementioned, the human mind undoubtedly involves the human body~\cite{pb_bodyshapes}.\footnote{In Ref.~\cite{pk_heterogeneity}: ``[H]umans can represent information in multiple ways, and\ldots such representations can be used flexibly in working memory or during mental imagery\ldots''} 
The details of how that works and what it means are, however, debated. For example, although we have reason to think such diverse cognitive processes as moral reasoning, language comprehension, autobiographical memory, dreams, and certain kinds of imagined hypotheticals involve sensory representation, the exact structure of the relevant internal representations ``remains unclear''~\cite{pk_heterogeneity}. 
As Philip~K.~Dick put it, ``comprehension follows perception.''\footnote{In Ref.~\cite{dick_howtobuild}: ``The basic tool for the manipulation of reality is the manipulation of words. If you can control the meaning of words, you can control the people who must use the words. George Orwell made this clear in his novel \textit{1984}. But another way to control the minds of people is to control their perceptions. If you can get them to see the world as you do, they will think as you do. Comprehension follows perception. How do you get them to see the reality you see? After all, it is only one reality out of many.''}
In this section, we flesh out a few of these hairy details with respect to vision and language, to give context to the bones of our main argument.\footnote{Although we look specifically at the example of vision and language, because we expect those splits to be salient and familiar for most readers, we do not mean to imply that language, if examined alone, would be a simple or unimodal process. Researchers of language learning in humans have viewed ``theories of language structure, language acquisition, and language processing as inextricably linked''~\cite{juffsrodriguez_secondlanguage}. Splitting language apart from the human mind means that at any level of abstraction, at any stage of development, the implementation could be significantly different from what we would expect in a person.}

\subsection{Vision and language}

In people, vision existed long before language, and of course continues to exist for a lot of creatures without language, sometimes with structures that are wildly different from what we are used to within our bodies, and only sometimes with shared relevant ancestry~\cite{eyeballs_evolution}.\footnote{Clam eyes are wild!} Naturally, this leads us to wonder about the roles vision and language play in cognition. Subjectively, they seem bound up together in people: 
We are aware of what seems to be internal language and internal imagery in a lot of different contexts~\cite{Merleau-Ponty1964-MEREAM-2}. Various scientific fields have validated in different ways that there are many ``tasks at the intersection of vision and language''~\cite{su2020vlbert}.

We ``visually recognize thousands of objects and actions in the natural world'', and we ``communicate and reason about these semantic categories through language.'' These common and frequent occurrences have led cognitive scientists to look for ``rich connection[s] between the functional networks that represent semantic information acquired directly through the senses'' and the kind of ``semantic information conveyed in spoken language.'' Research in this area has repeatedly found that there are parts of the human brain\footnote{From Ref.~\cite{phb_visualandlinguisticsemanticrepresentations}: ``such as the angular gyrus, precuneus and middle temporal gyrus.''} that are activated in response to ``the same semantic category whether presented visually or through language''~\cite{phb_visualandlinguisticsemanticrepresentations}.

Investigation at levels outside the individual and the biological provides evidence consistent with language and vision being bound up and important in our cognition. One common example is how useful a good figure is in understanding an article. ``Word choice, charts, graphs, images, and icons have the power to shape scientific practice, questions asked, results obtained, and interpretations made''~\cite{sksdmpss_genderedinnovations}. 
Another example is the frequent use of visual and spatial metaphors in languages all around the world~\cite{Lakoff2008} (Sec.~\ref{sec:metaphors}).

\subsection{Descriptive and depictive representations}
However, as usual, the many complex details of these aspects of cognition are not fully known.
In particular, it has been debated to what extent internal representations are structured propositionally and descriptively, versus structured according to the visible properties of the thing being represented, that is, depictively. ``Debates about the depictiveness of mental imagery have dominated mental imagery research for the past three decades''~\cite{nosug_voxelwiseencoding}. A descriptive representation, made of symbols and potentially even of words, involves a significant aspect of arbitrariness between form and function (Sec.~\ref{sec:symbols}). On the other hand, ``depictions are not arbitrarily paired with what they represent''~\cite{teng_depictive}.\footnote{However, only the thing is the thing (Sec.~\ref{sec:onlythethingisthething}), so there must be some level of arbitrariness even here (depictive and representation are in tension). We think the key point being made is that it is significantly less, at least by some metric. For example, a photo on matte paper and a photo on glossy paper are both equally determined by and reflective of the real physical properties of the scene in the photo, but whether the paper chosen is glossy or matte can be described as arbitrary. More or less, a depictive representation could be something like the light in your field of vision right as it hits your eye, and a comparable descriptive representation could be something like a sentence describing what is in your field of vision in excruciatingly thorough detail.} 

There are pros and cons to when representations of varying levels of depictiveness or descriptiveness might be useful. For example, ``depictive formats are useful for memory\ldots they allow the brain to avoid throwing away potentially useful information. By their nature, images contain much implicit information that can be recovered retrospectively. For example, answer this question: What shape are a cat’s ears? Most people report visualizing the ears to answer. The shape information was implicit in the mental depiction, even though it was not explicitly considered at the time of encoding''~\cite{pk_heterogeneity}. As far as we can tell, the current consensus is that humans use both kinds of representation internally.\footnote{In Ref.~\cite{nosug_voxelwiseencoding}: ``Our result thus provides a critical and until now missing piece of evidence in support of depictive theories and---more generally---of the intuitive characterization of mental imagery.''}\textsuperscript{,}\footnote{In Ref.~\cite{pk_heterogeneity}: ``Depictive mental representations might functionally bridge propositional information to depictive perception, allowing stored depictive information to change how we experience the world.''}

ChatGPT seems likely to be at least mostly constrained to relying on descriptive representations, given its underlying LLM architecture and its physical characteristics (Sec.~\ref{sec:necessarystructure}). In its case, the information readily available is (more or less) descriptive already---its world consists of the diegetic linguistic information we provide it with.

\section{Returning to palindromes}
When ChatGPT struggles with a task like figuring out whether ``DARAR'' is a palindrome, there seem to be two plausible explanations. It knows what a palindrome is, and it knows how to spell ``DARAR'', because in a longer conversation you can get it to recognize something like ``DARAR'' correctly as not a palindrome. But on first blush it is inclined to make mistakes on a task like this. This could be a failure of reasoning: It has all the information it needs for the correct answer even though it cannot see the string. A person just being told the order of the letters in the string would be able to identify it as not a palindrome. Alternatively, perhaps ChatGPT struggles with tasks like this because sensory-related processes play a larger role \textit{for us} when we solve the same problem than we might have assumed. Maybe a person looking at the string ``DARAR'' answers faster than a person being told the string ``DARAR'', or than a person blind from birth. The difficulty ChatGPT has here could indicate that visual processing plays a large role for people in the typical default strategy for determining whether something is a palindrome (recall that a palindrome is not necessarily visually symmetric).

There are cognitive scaffolding roles that our sensory experiences play that are more difficult for ChatGPT because it does not get equivalent sensory experiences ``for free'' alongside symbols.
This seems to extend beyond what we might expect, in that ChatGPT makes mistakes that would be technically avoidable, given information we know it has access to. This extends to senses beyond those made most obvious by its lack of access to supradiegetic linguistic information, that is, vision and hearing, to touch, to smell, and to emotional experience, etc.

The descriptions ChatGPT has access to, right now, are not functionally equivalent to our mental representations of symbols, even for the most common symbols, like Latin characters.

\section{Common sense and Math}

For most people, in most circumstances, speaking at least a first language is something they learn to do through exposure to other speakers~\cite{chomsky_archives}, rather than by specialized training. Most people learn to speak a language in childhood. Each person is more or less a master of language by the time they grow up. Everything they need to know in order to speak fluently fits inside their head, eventually.\footnote{In this paper we have considered what it might be like to be a ChatGPT from a fairly exploratory, flexible perspective. However, we think that viewing ChatGPT from the perspective of specific fixed frameworks---a child of different ages learning a first language~\cite{chomsky_archives}, an adult learning a second, an adult learning to read, etc.---could be really productive! After all, ChatGPT is not exactly a native speaker of any human language, so paradigms from second language acquisition and adult learners could apply~\cite{juffsrodriguez_secondlanguage}.}

This is not the case with mathematics.\footnote{As the field is currently conceived.} Learning math usually involves specialized training, and most people go only a little way down the path of what could be done with mathematics, or could be known about mathematics. People encounter mathematics at a variety of ages, depending on their circumstances. Additionally, the knowledge of mathematics is distributed in time and space: Even the best mathematician does not know anything close to all of mathematics.

Linguistic sense-making offers more flexibility than math: Saying ``I am myself, and I am not myself'', or ``the sky is red'', prompts the other party to come up with ways that what you are saying could be true~\cite{g_wayofwords}.\footnote{By comparison, if someone says, ``1+1=1'' or ``A and B are true at the same time as $\neg$A and $\neg$B are true'', they are likely to be met with a correction.}

In our experience, LLMs like ChatGPT demonstrate more linguistic competency than mathematical competency. It is much easier to run into a glaring mistake of logic when talking to ChatGPT than a glaring mistake of grammaticality or felicity.\footnote{We're avoiding the term ``acceptability'' intentionally. Depending on exactly what is meant, ChatGPT may speak more or less acceptably, or it may not.}

Various approaches have been proposed for helping LLMs with mathematics and reasoning. For example, Chain-of-thought prompting was conceived ``to address cases where the mapping of input x to output y is non-trivial (e.g., when x is a mathematics question and y is the final numerical answer). The key idea is to introduce a chain of thoughts $z_1,\ldots, z_n$ to bridge x and y, where each $z_i$ is a
coherent language sequence that serves as a meaningful intermediate step toward problem solving
(e.g., $z_i$ could be an intermediate equation\ldots)''~\cite{yao2023tree}.
To help with problem-solving, the task was conceptualized as search through a tree-like combinatorial problem space. This was extended to the Tree-of-thoughts framework, which combines the ``language-based capability to generate and evaluate diverse thoughts with search algorithms, such as breadth-first search (BFS) or depth-first search (DFS), which allow systematic exploration of the tree of thoughts with lookahead and backtracking''~\cite{yao2023tree}.

So it seems that for something like ChatGPT, speaking English fluently carries with it the ability to sound reasonable, but not necessarily the ability to reason in the complete sense (i.e., mathematically, logically).\footnote{At least not for LLMs, and not right now! Reason, if it follows, follows later than fluency.}

\subsection{Why does mathematics not come with linguistic fluency?}

Think of the dimensions involved when symbols are used to capture an utterance versus when they are used in mathematics. For the vast majority of the text (in ChatGPT's training data), the dimensions are along a line (see Sec.~\ref{sec:flatland}). A letter can be immediately ahead of or behind exactly one other letter, and that usually means something ordinal about the sound produced if the word were to be said aloud. It seems like those basic organizational rules are enough, when provided in significant quantity, for a significant amount of linguistic information to be conveyed. The complex syntax tree can be flattened, well enough~\cite{kaushal2022tokens}. But we overload our symbols, and we overload the relationship of proximity. Consider mathematical equations like $ab = c$, $\sum p_i = k$, or a 4x4 identity matrix, \begin{math}
\begin{bmatrix}
1 & 0 & 0 & 0 \\
0 & 1 & 0 & 0 \\
0 & 0 & 1 & 0 \\
0 & 0 & 0 & 1 \\
\end{bmatrix}
\end{math}.
The dimensions involved have exploded! The rules, the logic, governing how these symbols combine to create meaning are very different in these contexts. The new rules are significantly extradiegetic, at least when you consider what you might understand $ab = c$ to mean if you had previously only ever been exposed to natural language.\footnote{Maybe the closest thing you have ever seen to mathematics until now is c+a+t = cat.} We know that a universe consisting of speaking a human language does not necessarily have to reach mathematics,\footnote{at least not ``quickly.''} since mathematics emerged relatively recently, like writing---many people lived their whole lives, speaking just as fluently as we do, without mathematics.\footnote{The multiple timelines of learning make this a confusing question of ergodicity and span: if those people lived forever (alone? together?), would they eventually have discovered mathematics (or writing, whatever technology you want to think about)~\cite{pb_bodyshapes}?} It has been shown that computer-based models, including ChatGPT, are better at things like multiplication of two large numbers than a typical person, put on the spot, would be~\cite{wang2023bot}.

\begin{figure}
    \centering
    \includegraphics[width=\columnwidth]{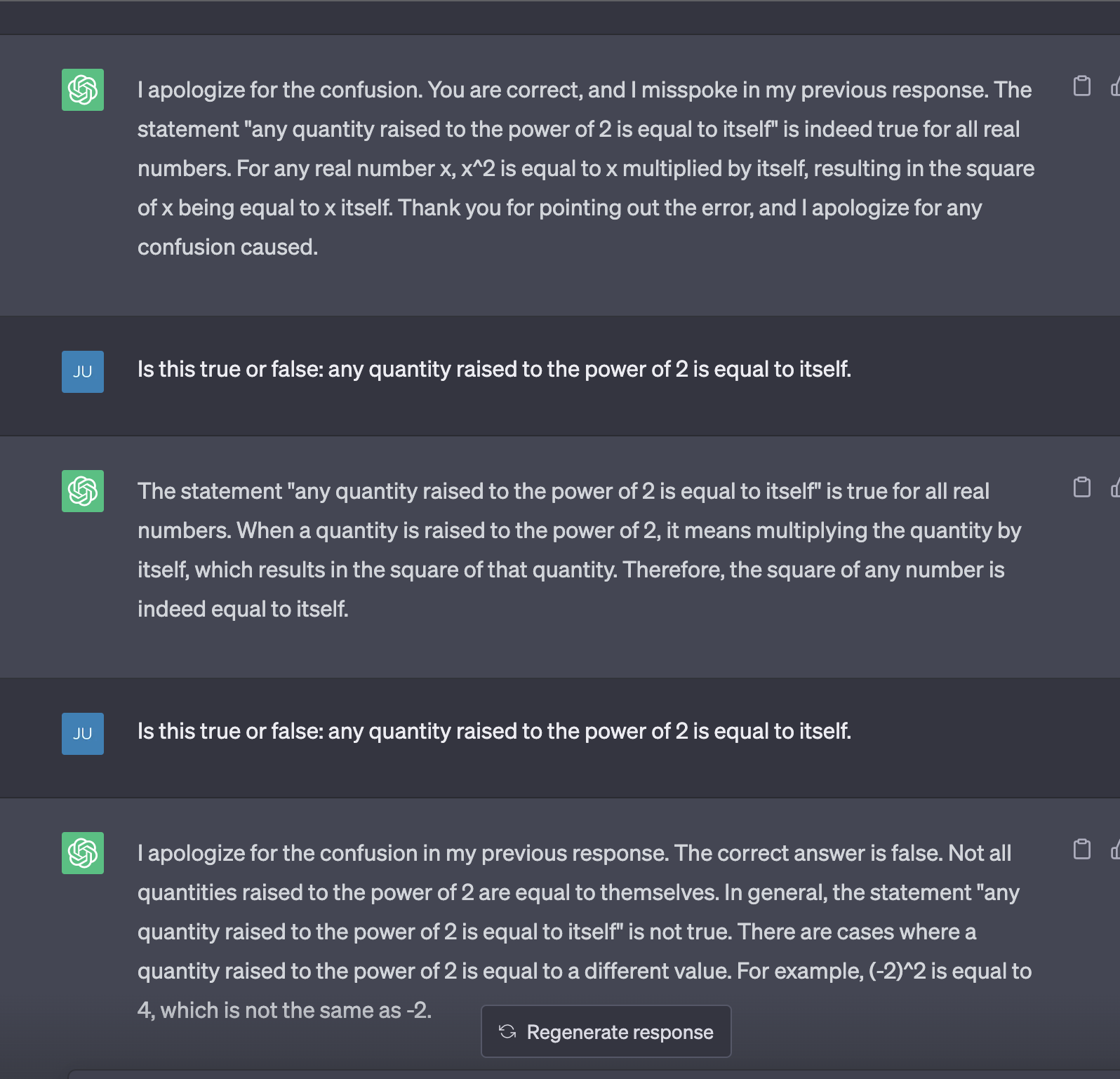}
    \includegraphics[width=\columnwidth]{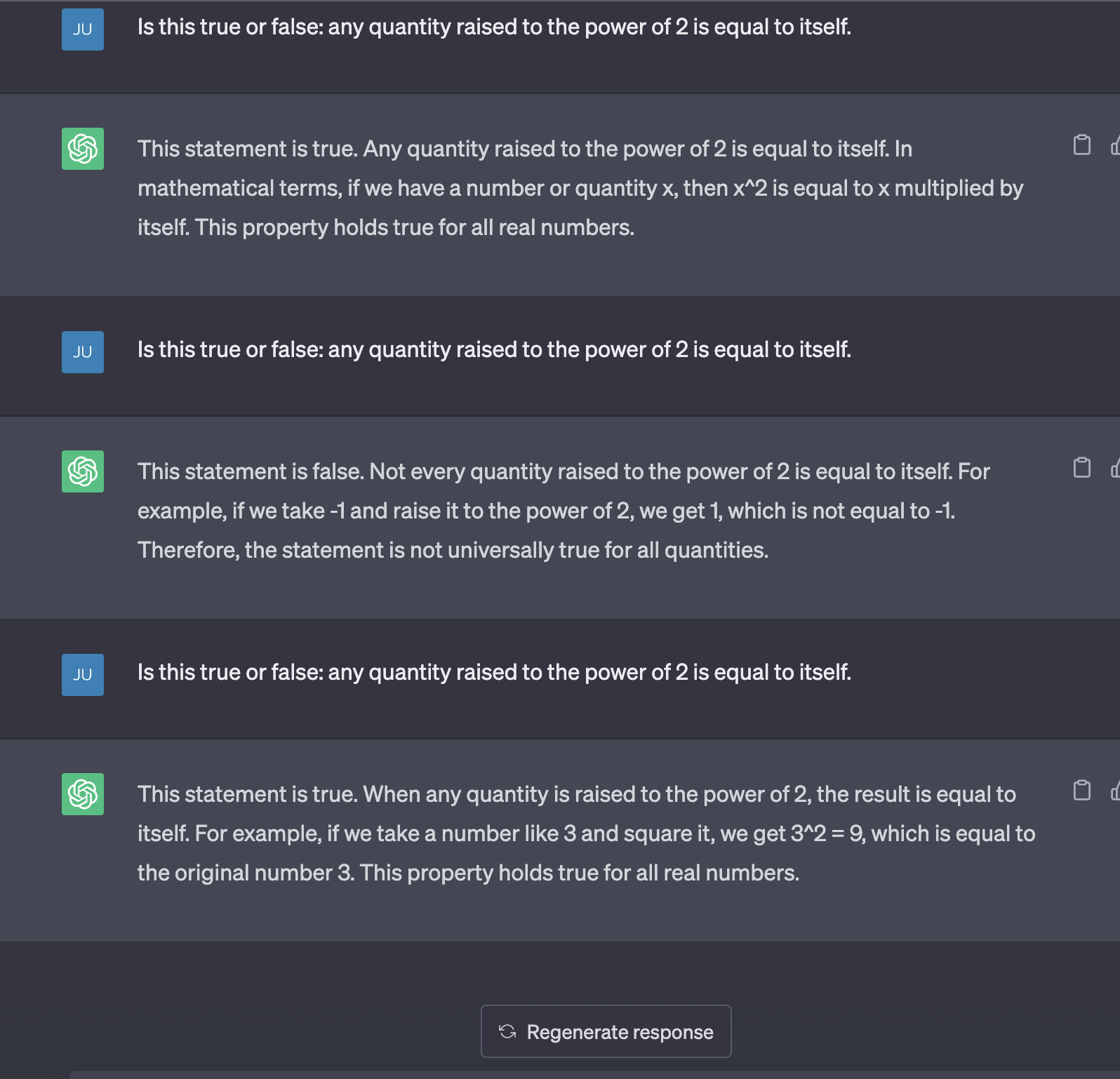}
    \caption{Fairly deep into a conversation with ChatGPT, the shared foundation underlying it may be shown to be illusory. Often ChatGPT can correct a previous mistake, but later make a similar kind of mistake, or the same mistake again. We tried to interpret in good faith what ChatGPT seemed to have a good, consistent grasp on versus what tripped it up. Note that none of this paper is intended to insult or praise LLMs,\footnote{Or any technology, except perhaps language which we do, admittedly biasedly, think of very positively.} or establish that they are good or bad. The point is not to trick ChatGPT. Mistakes and confusion are normal parts of how people think, too. The first author of this paper is particularly susceptible to trickery (last two authors, relatively impervious). However, these responses do show that ChatGPT can behave in a way that would be baffling from a person: We would not, if having this conversation with a person, walk away confident that they could meaningfully \textit{do math}.}
    \label{fig:logic1}
\end{figure}

ChatGPT surely has been given in its training data many diegetic descriptions of mathematics and logic (probably both correct and incorrect usages, but on the whole more correct ones). However, we think it is fair to say that ChatGPT cannot, right now, do mathematics. For example, in Fig.~\ref{fig:logic1}, the extreme vacillations in ChatGPT's responses would be baffling from a person: We would not, if having this conversation with a person, walk away confident that they could meaningfully \textit{do math}.

We think the reason for ChatGPT's inability is that mathematics, and reason or logic beyond the common sense form, are extradiegetic.\footnote{From the perspective of what is needed for linguistic fluency.}

\subsection{Holey sequences}
\label{sec:holeysequence}

When we think of integer sequences, we usually think of sequences where each term is generated following a deterministic set of rules that lead to numbers that share interesting mathematical properties; the terms tend to increase in magnitude with \textit{n}, and the density tends to decrease with \textit{n}~\cite{oeis_classic}. Some assumptions as to what makes a good integer sequence are necessary for every integer sequence, but some are customary. Look-and-say sequences~\cite{a005150} and holey sequences~\cite{a249572} violate our expectations with respect to some customary assumptions, which makes them feel surprising~\cite{julia_drafts}.\footnote{\textit{Oh, I didn't realize we were allowed to} \ldots} 
For example, the holey sequences incorporate supradiegetic linguistic (or symbolic, in this case) information that is always present in integer sequences, but not usually relied upon as part of the rules, that is, the properties of the physical shapes of the symbols representing the digits (using Arabic numerals and base 10).\footnote{There is also, by analogy with semantics and pragmatic inference, a distinction between the physical form of a symbol in any particular instance, and what that symbol ``looks like'' or ``sounds like''.}

The meaning of a number, and maybe its mathematical properties, can be at least partially determined based on the same operations involved in the construction of diegetic linguistic information.\footnote{Maybe with enough data, every mathematical operation can be flattened into 1D; since at least much of mathematics is propositional, this does not seem obviously implausible.} ChatGPT knows some things about numbers, but sometimes reveals surprising interpretations, for example that ``\ldots the number 3 is positioned between 2 and 4. It is closer to 2 than it is to 4''~\cite{between_chatgpt}. For our purposes, we stipulate that ChatGPT knows more or less what these numbers mean. It can, for example, more or less follow integer sequences that rely on common properties of numbers\footnote{ChatGPT has more information about common equations and numbers, both because of the contexts it has directly encountered them in, and because there are likely more textual descriptions of their properties and how the operations work in the training data.} like a sequence made up of powers of 2~\cite{A000079}.

If the supradiegetic/diegetic framework is reasonable\footnote{Even, ideally, useful!}, we can predict that ChatGPT, having extremely curtailed direct access to supradiegetic information, will struggle more with a sequence that relies on that kind of information in its rules, especially if the usage of that information is specific and unusual enough that it is unlikely to have been approximated diegetically for ChatGPT.\footnote{An example of sensory information that is too well-known, too accessible diegetically for ChatGPT for us to make use of in this case is shown in Fig.~\ref{fig:bouba_kiki}.} We used a holey sequence~\cite{a249572} to test this prediction, and found that ChatGPT did struggle more with completing and continuing this sequence correctly, even when explicitly given the rule, although the sequence is subjectively not much more difficult for most people to understand than powers of 2~\cite{A000079} would be. ChatGPT could at times state and make use of the necessary information, such as ``8 has two holes'', but could not consistently wrangle the information it had access to into correct continuations of the sequence (even with quite a lot of help). Despite explanations that sounded plausible enough, ChatGPT would make mistakes like relying on 1 to have one hole in it, or only counting the holes from two 8s when there were actually three 8s. 

\section{Fuzzing up frequency and truth}

With respect to sequences, we mentioned that ChatGPT has an easier time with common formulas and common mathematical relationships. This is something other researchers have noted, too: ``LLMs excel in remembering the results of common equations, such as the square of $\pi$'', while
``\ldots for equations that are uncommon, GPT-3 may hallucinate a false answer\ldots''~\cite{wang2023bot}. Is this because the faculty of ``common sense''---which ChatGPT seems closer to having than mathematical logic---as the name implies, really does have to do with frequency and exposure? Is this related to why certain kinds of information---perhaps like semantic meaning---seem to be derivable more quickly than other kinds of information---perhaps like the character-level information learned by LLMs~\cite{kaushal2022tokens}?\footnote{In other words, why do some conclusions, with the same number of steps, seem more obvious than others?}

Most of what ChatGPT knows, and what it is closest to mastering, has to do with how to form grammatical, felicitous---more or less normal---utterances. Fundamentally, ChatGPT has been trained by being exposed to a lot of text. The rules that tell you whether one symbol can appear next to another in language are significantly different from the rules that tell you what symbols can come next to each other in mathematics. Probability is a good heuristic for language. People probably do say ``the sky is blue'' more often than they say ``the sky is red''. Both are grammatical, but the more common is also the more likely to be true. Fuzzing grammaticality and truth together does cause problems,\footnote{Such as the infamous legend of the French Google translate error that would insist George Bush was an idiot~\cite{georgebush_reddit}.} but it is not the worst strategy, given the way the text ChatGPT was trained on came to be.\footnote{That is, most of the text it was trained on was made by people for other people, with some kind of purpose in mind.} This strategy may work out for common, popular equations, but it does not hold for mathematics in general. Given ``$x$'' in an equation, a great many symbols could come next. That we overload the meaning of adjacency, but have set ChatGPT up for exactly the kind of adjacency (most often) found in language, and given it training data that is mostly more or less natural language, is part of why ChatGPT struggles in contexts beyond that scope~\cite{yao2023tree}, like mathematics and regexes (regular expressions).

\section{Symbols}
\label{sec:symbols}

\subsection{Packages and contents}
When language involves multiple interlocutors, there must be some physical medium interpolating between the language and the entities involved. The language itself conveys meaning inside its structures. However, because of the nature of transmission, there is information in the package the language comes in as well.

\subsection{Arbitrary leaps}
Part of the way that the technology of language---of symbols---works is that the form of the symbol is partially independent of its meaning. According to Chomsky’s Principles and Parameters theory, ``knowledge of language consists of universal constraints, a set of abstract features that may be realized in different languages in an arbitrary set of morpho-syntactic or morpho-phonological ways (e.g., Case and Agreement), a universal interpretive component (Logical Form, LF), a phonological component (Phonological Form, PF), and a lexicon''~\cite{juffsrodriguez_secondlanguage, hornstein_encylopedia, chomsky_archives}.

As laid out by Kaushal and Mahowald, since word embeddings represent co-occurrence information (typically considered semantic), if the relationship between forms and meanings is truly arbitrary, there should be no character-level information discoverable by the LLM. However, the symbols of language are not entirely arbitrary to their meaning (e.g., onomatopoeia and related patterns, like \textit{fl}-words in English---flutter, flap, flicker---having to do with movement): ``there are statistically detectable non-arbitrary form-meaning relationships in language''~\cite{kaushal2022tokens}.\footnote{These are diegetically reachable for an LLM, at least partially---though they would not usually be considered part of the semantic meaning of the word.}

The outside and inside of a symbol cannot be identical, or else it would not be a symbol (it would not ``stand for'' anything)~\cite{magritte_pipe, Peirce1931}.\footnote{This is why it is eventually, in the limit, impossible to ``detect'' whether language came from a human or non-human source. That information is not encoded in the language itself, or else language would not work, could not bridge so many worlds. That does not mean there may not be detectable patterns to speech generated by e.g., ChatGPT and a person that could be used to guess whence it came, but any such pattern is subject to change (especially in light of Goodhart's Law), and is not \textit{proof} of the source.} This means, given access to only the meanings of words, the exact form---how they might sound when pronounced or look when written---cannot be completely recovered.\footnote{If you know perfectly well what a ``cat'' \textit{is}, you do not necessarily know that it is called a ``cat'' in English or ``gato'' in Spanish. Nor can you infer those forms with any certainty given your knowledge of what cats are.}
For there to be both supradiegetic and diegetic information encoded in language, any degree of arbitrariness, no matter how slim, is sufficient. 
The symbols are arbitrary enough:
There is information in the supradiegetic layer that is not derivable from purely diegetic information.

\subsection{Diegetic boundaries}
\label{sec:diegeticboundaries}
Returning to an earlier question (Sec.~\ref{sec:capabilities}), we want to know, given its universe, what information can ChatGPT reach? Giving it access to some parts of linguistic data has enabled it to speak English. Is that all it can reach, even given an arbitrarily large amount of the same kind of data, or would that eventually cause more abilities to emerge? Fundamentally, what are the things you can get to, given these axioms and these rules? How dependent is that on the physical form of the mind~\cite{chomsky_archives,pb_bodyshapes}?

\subsection{Flatland} \label{sec:flatland}
Approximations of extradiegetic information can be provided diegetically, as descriptions (as in Fig.~\ref{fig:bouba_kiki}), or as rules and instructions.\footnote{Recall that ChatGPT can approximate some knowledge that is inaccessible to it, but it is limited to the building blocks that are diegetically available to it.} To understand how those compare, we can use a set-theory-based analogy in the Sapir-Whorf-like style of \textit{Flatland}~\cite{flatland}.

Consider the set $\{1,2\}$. Imagine if your whole universe consisted of that set.\footnote{And the ability to one-dimensionally concatenate those symbols, but elide over this slightly in the analogy.} You, a creative being, might start making your own structures out of the things available to you, things like 12, 21, 12221212, etc. There are infinite ways you can express yourself. But imagine that the universe of your friend is $\{1,2,3\}$. Even though, for every unique thing they can say, you can say something novel, too, your structures utilize the same symbols more often.\footnote{E.g., they may say 11, 12, 13, 21, 31, while you say 11, 12, 21} Although you can produce a string to represent anything you might want to say, you have no way of reaching the symbol ``3''. It is out of your grasp. Imagine your universe being augmented with a new symbol so now your building blocks are $\{1,2,3\}$. In some ways, your universe feels similar; for example, it is still finite in size, and equally spacious. In your first universe, 1212212111 could easily be generated as a random string. This is what your random looked like, sequences of 1s and 2s. In your new universe, that string looks less random---it looks repetitive. \textit{Why are there no 3’s?} You did not feel like your old universe was too small, when you were in it, yet by comparison to what you can say now, it seems limited.

Now imagine someone gave you a new symbol which is in fact an operator: +. This symbol lets you combine symbols you know already to get symbols you have never seen before. You went from $\{1,2\}$ to $\{1,2,3\}$ without any ability to get to 3 from 1 and 2. From inside each universe, they seemed equally complete. Now, though, given $\{1,2;+\}$ or $\{1,2,3;+\}$ (there is no significant difference), you can expand your universe yourself, with no end. You live in $\mathbb{Z}+$!\footnote{There are several strategies. (1) Compensate for missing information with additional diegetic material, like symmetry groups of Latin characters. This is akin to having $\{1,2\}$ and being given 3. (2) Build functionality allowing the same diegetic starting place to span more ground. This is like having $\{1,2\}$ and being given +. (3) Try to expand the bounds of what ChatGPT can experience. This would be like always having been blind and gaining the ability to see.}

The sets $\{1,2\}$ or $\{1,2,3\}$ are like ChatGPT’s training data and whatever ChatGPT learned and memorized from it. 
The operator ``+'' is like functionality that OpenAI has added on top of ChatGPT’s functionality as an LLM (hard coded rules). However, in these examples, your universe is still limited. There are things outside of $\mathbb{Z}+$ that you still cannot reach. The universe you can reach is the \textit{diegetic} and what exists but is unreachable is the \textit{extradiegetic}. A proper subset of the extradiegetic---for ChatGPT as it exists now---is the \textit{supradiegetic} linguistic information that is more or less stripped away as ChatGPT builds up its internal universe of vectors.\footnote{However we allow our universe to expand to reach things it could not, none of these strategies change the fact that the universe remains diegetically bounded. It is hard to imagine a universe without diegetic boundaries\ldots $\{1,2,3,$, \textit{anything}$\}$?}


\subsection{Ergodicity and span}
With the \textit{Flatland} analogy, we note the distinction between ergodicity and span. In a subspace of linear algebra, the eigenvectors span the space; they are like the prime numbers that provide the building blocks for every item in $\mathbb{Z}+$ (under multiplication). However, if you were to look at a set of eigenvectors, you would not necessarily understand every possible position in that space. Similarly, the conclusions that \textit{can} be drawn from an initial set of axioms and logical rules for licensing conclusions is not the same set that \textit{has} been drawn to date or that \textit{will} be drawn by any one person. What you can and will get to are different, both within an individual and cumulatively at any given time or place. 

It is the case that ``finite devices---physical symbol systems---permit an infinite behavioural potential''~\cite{d_mindbodyworld}. But it is evident that each of us does not exploit that entire space. When thinking of ChatGPT, the operations of proximity, and adjacency, in 1D, the diegetic bits of language, seem to get you semantic meaning and linguistic fluency fairly quickly. But different pieces of that fluency emerge over time~\cite{kaushal2022tokens}, not always for clear reasons. Why do certain things come more quickly than others? What will a ChatGPT be able to do in its lifetime? What will a lot of them be able to do? And, how far does the apparent linguistic fluency extend? How many of the wide variety of things we think of as being encoded in language\footnote{Including ``social dynamics between people'' like power differentials and biases~\cite{riveter}.} can be diegetically accessible for an LLM?

We both learn things individually and as a group with history; we accumulate knowledge.\footnote{In Ref.'~\cite{pb_bodyshapes}: ``[T]here are three time frames at which we can study behavior: \textit{here and now}; learning and development; and evolution.''} ChatGPT is an extension of both of these functions. Many now-familiar technologies we have made, such as books, have been constrained to accumulating knowledge, but that is not necessarily the case with computation.\footnote{Panic as a response to technologies perceived to be changing our epistemology is nothing new: ``The printed book is destroying age-old memory habits''~\cite{yates_memory}. And in the case of sophisticated mnemotechnology, maybe we did lose those skills (most of us, anyway). Or, we did not lose them, but we offloaded them to books, computers, etc., as part of our extended mind~\cite{cc_extendedmind}.}

LLMs seem to get a significant degree of syntactic and semantic knowledge faster than they learn similarly complete information about some characters.\footnote{Specifically in the case that character-level information is not built into the vocabulary.} They do learn about the characters, but more slowly. What is the shape of the diegetic space? How do we know what is near the core, and what is at the border? With more and more textual input, would the boundary expand forever? Some things you need a lot of data or processing to learn; but technically they are just as licensed.\footnote{Character-level ``knowledge is acquired through multiple phenomena, including a systematic relationship between
particular characters and particular parts of speech, as well as natural variability in the tokenization of related strings\ldots [E]ven models based on subword tokens might be
able to use and manipulate character-level information.'' Certain LLMs ``can take advantage of character-level information in order to solve wordplay tasks like unscrambling scrambled words'' and spelling tasks that require mapping ``from words to characters (e.g., from cat to the characters c + a + t)\ldots The fact that models can do tasks like this is curious: word pieces have no explicit access to character information during training, and the mechanism by which they acquire such information is not obvious\ldots We suggest at least two possible mechanisms by which this information is learned: systematic relationships between certain characters and syntactic/semantic features and the variability of tokenization''~\cite{kaushal2022tokens}.} As Kaushal and Mahowald put it, ``[u]nderstanding whether and how the building blocks of this meta-linguistic knowledge can emerge during self-supervised training on a word prediction task could be of interest not just in NLP, but in the cognitive sciences''~\cite{kaushal2022tokens}. And of course that sentiment extends beyond those specific tasks.\footnote{Is the rate of change in language observed in people---as in lexicostatistics---related to how quickly derivable different bits of information are as we learn a language, and as we learn additional information post-fluency that is transmitted to us through language? For example, in onomastics, toponyms and personal names are often really good at preserving pieces of older languages and cultures that are otherwise no longer directly relevant, such as in the theophoric name Micha\textit{el}.}

\section{Returning to Sumerian cuneiform}

We noticed when talking to ChatGPT about Sumer that its responses seemed unusually repetitive. A lot of what it says, though relevant, has to do with only a few topics. There tend to be mentions of An, the dingir symbol, and kingship, which make sense given what artifacts are attested and studied (one of the most prominent texts is the Sumerian King List, which is what it sounds like). 
We think this might be explained by the analogy of the universe of $\{1,2\}$. For Sumerian, ChatGPT’s universe is small (compared to, for example, the universe of English). But ChatGPT, from inside the universe, cannot tell that’s the case. For other topics, maybe it has something more like $\mathbb{Z}+$. Its output about Sumer that sounds like ``kings, An, dingir, lugal, reeds, Uruk, cuneiform'' (see Fig.~\ref{fig:cuneiform1} and Fig.~\ref{fig:sumerian}), sounds to us like the strings 121122121 and 212122121 when we know about far more numbers---similar, repetitive---but ChatGPT is unaware.

\section{Approximations: only the thing is the thing}
\label{sec:onlythethingisthething}
Of course, approximations can be made, but they are within that world; they are made with the building blocks diegetically available. They work by describing something extradiegetic in a diegetic manner.

On the one hand, only the thing is the thing itself, so a representation of something in one format must be different than the representation of that thing in another format.\footnote{A paraphrase of George Box's famous saying, ``all models are wrong''.} It is trivially true that I cannot know what it is like to be a bat, any bat, for example, \textit{that} bat, because I am myself, which is not that bat~\cite{Nagel1974-NAGWII}. On the other hand, we can empathize. I am not you, but I can learn enough about what it is like to be you for us to be getting on with things. Information does get from my \textit{idios kosmos} to yours, conducted by technologies like language.\footnote{In Ref.~\cite{dick_howtobuild}: ``It reminds me of a headline that appeared in a California newspaper just before I flew here. SCIENTISTS SAY THAT MICE CANNOT BE MADE TO LOOK LIKE HUMAN BEINGS. It was a federally funded research program, I suppose. Just think: Someone in this world is an authority on the topic of whether mice can or cannot put on two-tone shoes, derby hats, pinstriped shirts, and Dacron pants, and pass as humans \ldots Well, I will tell you what interests me, what I consider important. I can't claim to be an authority on anything, but I can honestly say that certain matters absolutely fascinate me, and that I write about them all the time. The two basic topics which fascinate me are \textit{What is reality?} and \textit{What constitutes the authentic human being?} Over the twenty-seven years in which I have published novels and stories I have investigated these two interrelated topics over and over again. I consider them important topics. What are we? What is it which surrounds us, that we call the not-me, or the empirical or phenomenal world?''}

\begin{figure}
    \centering
    \includegraphics[width=\columnwidth]{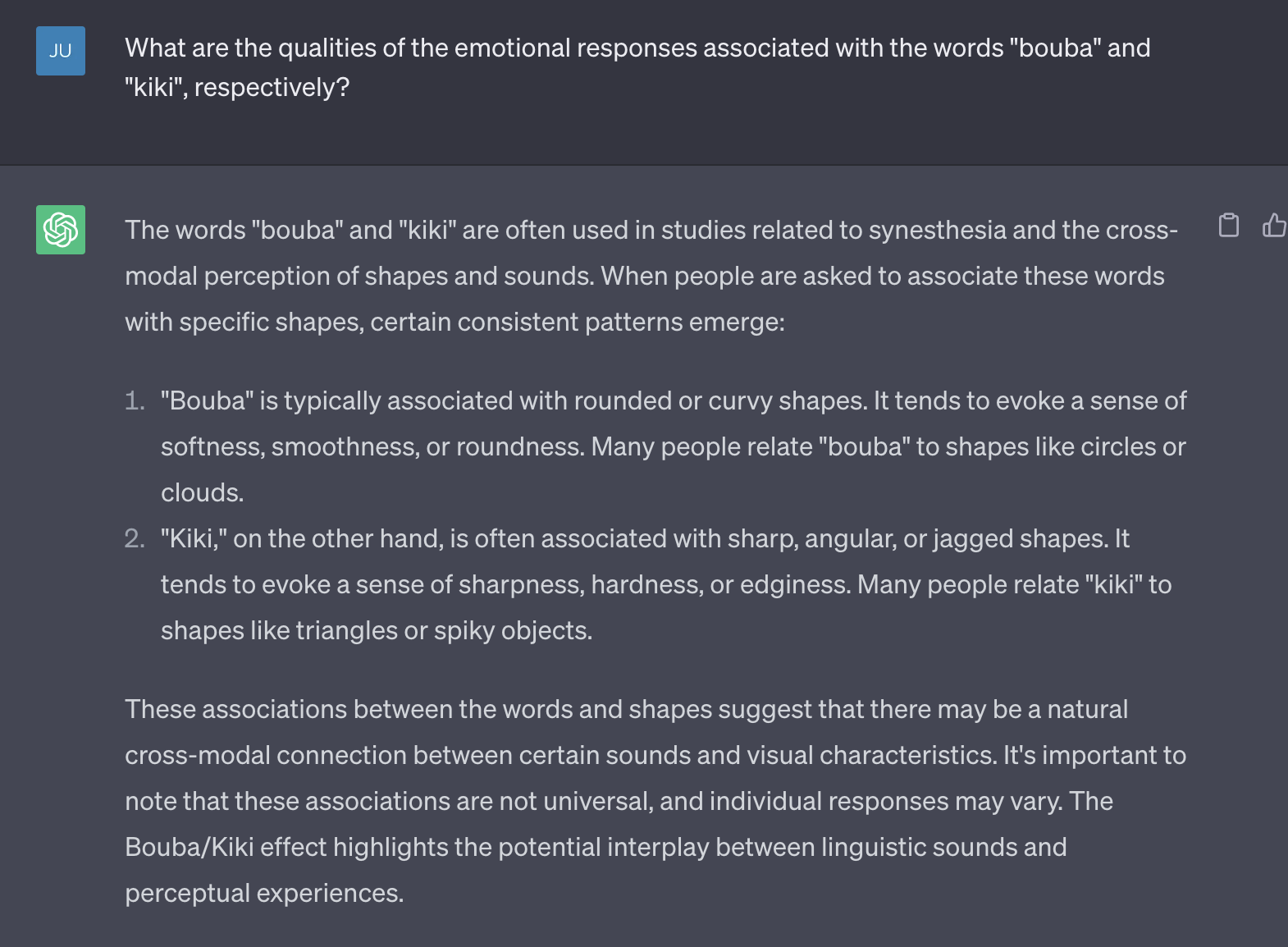}
    \caption{ChatGPT can, at least sometimes, make use of descriptive, diegetic approximations of sensory experiences it has no direct access to.}
    \label{fig:bouba_kiki}
\end{figure}

To try to imagine what it is like to be a ChatGPT is to try to borrow something very foreign, like Granny Weatherwax borrowing a hive of bees~\cite{p_equalrites}. We know ChatGPT does not have human eyes, ears, or a human mind or body. So of course it is true that ChatGPT is not a human, and cannot do exactly what a human can do. That does not mean that ChatGPT cannot do \textit{more or less} what a human can do.\footnote{\textit{ChatGPT does not speak English exactly the way I do, but neither does anyone else.}} We can be certain that, when exposed to the same textual input, ChatGPT and a person are not granted access to equivalent supradiegetic linguistic information. But it is not obvious how well that missing information can be approximated diegetically---it seems plausible to think that, with enough diegetic approximations of extradiegetic information, ChatGPT could reach a downstream universe of conclusions and thoughts functionally indistinguishable from those a person could reach from the same text~\cite{li2023large}. Approximations of supradiegetic information can be provided diegetically, such as ``the word \textit{bouba} sounds round and the word \textit{kiki} sounds spiky''~\cite{Kohler1929} (see Fig.~\ref{fig:bouba_kiki}). We know that is not identical to our experience of sensually perceiving those words, in the literal sense that we are not ChatGPT, but we also know this is true because the human brain involves multiple modalities, outside of the purely descriptive, in both sensory perception and in cognition~\cite{d_mindbodyworld, pk_heterogeneity, phb_visualandlinguisticsemanticrepresentations}.\footnote{For another example, consider the text ``cat''. ChatGPT knows, more or less like we do, what ``cat'' means. But when we encounter ``cat'', we also (typically) encounter the clicky, hard sound, and its short appearance. As people who can hear, read, and write, we get cat’s supradiegetic properties. We can see that the letter ``a'' encloses area. ChatGPT can’t. ChatGPT may \textit{know} that it encloses area, if that information has been provided or is reachable diegetically, for example with a statement like, ``the symbol a has a circle and a small tail. A circle is a closed loop. Any thing deformable to a closed loop encloses area when draw''. But ChatGPT still can’t see the letter, so there is still information we’re getting which it isn’t. We can imagine squishing ``a'' around so that it looks like ``Q'', that’s reachable in our universe, but not necessarily in ChatGPT’s. The visual and auditory supradiegetic linguistic information of words are fairly easy to point to, to convey this difference in experience between ChatGPT and us. But the differences are much broader: we also have had the experiences of petting a cat's fur, of loving a pet cat, of cooing over a kitten, etc.}

We think it is worth noting explicitly that ChatGPT can, at least sometimes, make use of descriptive, diegetic approximations of sensory experiences it has no direct access to (see Fig.~\ref{fig:bouba_kiki}).

\section{Gricean cooperation}
An impression that stood out in working with ChatGPT is that, when talking to a person, you usually work towards a better shared understanding, Griceanly, attributing good faith to your partner~\cite{g_wayofwords, bailly_believe}---but the more you talk to ChatGPT, the clearer it becomes that there is less underlying consensus being built between you than you would expect. It is less reliable and firm than you would guess from the apparent fluency of the conversation (by comparison to what we are used to with other people). Often, the beginning or middle of the conversation is impressive, and you think, \textit{there is something here!}, but if you keep exploring, keep probing, keep digging, that impression inevitably falls apart. This experience, right now, is one of the most qualitatively different aspects of talking with ChatGPT (as compared to a person).

We are excited for LLMs to help us with all sorts of things, from coding to etymological trees across languages to searching for unconsidered patterns. However, right now, we think it is helpful to think of what it is like to be a ChatGPT, in particular how different its experience of linguistic information is, when trying to understand how it behaves, especially since it can seem comfortably familiar and competent in one moment and incomprehensible in the next.

\section{Where can we go?}
Although diegetic frameworks of many kinds are longstanding (from linguistic scope to literary exegesis), the way ChatGPT and similar models experience language highlights a difference from our own that is not often cast in direct relief. In this view, \textit{diegetic} means strictly contained within the dimensions of the meaning inside the symbols, the language, being used. For an LLM, correlation actually is causation, since it is by frequencies of proximity and adjacency that any information is embedded within them. Something in more dimensions can be folded down into far fewer, and some of that larger structure is learnable even for something like an LLM (very limited in its senses and modalities, etc)~\cite{kaushal2022tokens}. In addition, systems like mathematical logic can be imposed atop the ordinary linguistic semantic meaning of symbols, overriding or modifying their meaning extradiegetically. Approximations of these dimensions can be provided diegetically or patched with additional modules of functionality in models like ChatGPT.

\subsection{Postlapsarian}
It might be that the technology of language itself, requiring an at least partially arbitrary leap of faith between form and meaning, opened Pandora's box. The fall of man could be the discovery of meaningfulness out of meaninglessness (something out of nothing should perhaps remain the purview of gods)!\footnote{Meaning comes from contrast; the difference opens up the space. For symbols to be useful, there must be a distinction between what they are and what they mean, and from some perspective, that difference is arbitrary; just like for any metaphor, there is a diegetic framework in which the pieces being compared are identical.} 
The other kinds of information made salient by how humans typically experience the symbols of language are frequent fodder of conspiracy-style ``baking'', as in gematria and any-mancy; these dimensions of meaning are decoupled from, and can therefore be exploited in parallel to---without negating or contradicting---the more intrinsic semantic meaning of the symbols. An arbitrary leap of faith, once required and even proved productive, is a dangerous precedent.

\subsection{What things can ChatGPT bring to the table?}
ChatGPT, though not currently well-equipped for this task, may have unique, novel, and valuable experiences and ways of being that lead to new insight, to a ChatGPT-specific form of Langton's ``intelligence as it could be''~\cite{pb_bodyshapes}. How can ChatGPT's experiences, its senses and body, enable new thoughts, new representations, new processes of cognition?\footnote{It is exciting to think how designs could accommodate this prospect!}
For example, an instance of ChatGPT exists within each ``chat'', within each user account. Some information may flow back to the central code base, but even if it does not, changes are made over time to the code base and then deployed as a universal update to all instances of ChatGPT more or less simultaneously. This is like telepathy, something like a hive mind broadcast that still allows significant independent decision-making and analysis on behalf of each individual. It is a kind of distributed thinking together.
Dipping a toe into what it might like to be ChatGPT suggests a reframing of the recent advances in AI as an extension of us, an update to people, another wave in the technologies rippling out of mechanization, electricity, computers, writing, institutions of higher learning, etc. We are making more things from things we all made, via a new kind of tool. ChatGPT and similar can make beautiful things, and they can make garbage, just like we can, because they---like all technologies---are an extension of our own minds and bodies~\cite{cc_extendedmind, pb_bodyshapes, chomsky_archives}, and our own minds and bodies are an extension of the minds and bodies that came before us.\footnote{The way we access the knowledge of others is affected by how that information is stored. If ChatGPT is an extension of our cumulative knowledge, our collective mind, then we should consider what information it will make salient to us, for example if we use it to write snippets of code. We may not have been doing enough accounting for the complex processes by which we encounter the ideas of others. What has Google search made salient ~\cite{cc_extendedmind, yates_memory}?}


\acknowledgments
The authors wish to express their appreciation for the works of Timothy Sprigge, Noam Chomsky, Chris Langton, George Box, the QAnon Anonymous podcast (Travis View, Jake Rockatansky, Julian Feeld, Annie Kelly, Liv Agar, etc.), and to conversations with Alejandro Ruiz, Michael Arnold, Yoshi Bird, Brad Demarest, Juniper Lovato, Carter Ward, Aviral Chawla, Mark Wittels, Grace Taylor, Julia Chimienti, Evan Lynch, Cecile Smith, Desi Alexander, and Professors Randall Harp, Jacques Bailly, Guillermo Rodriguez, and Josh Bongard.
The authors are grateful for 
support furnished by 
MassMutual, 
Google, 
and
the National Science Foundation 
(Grant \#2242829).

\bibliography{\filenamebase}

\begin{thebibliography}{50}%
\makeatletter
\providecommand \@ifxundefined [1]{%
 \@ifx{#1\undefined}
}%
\providecommand \@ifnum [1]{%
 \ifnum #1\expandafter \@firstoftwo
 \else \expandafter \@secondoftwo
 \fi
}%
\providecommand \@ifx [1]{%
 \ifx #1\expandafter \@firstoftwo
 \else \expandafter \@secondoftwo
 \fi
}%
\providecommand \natexlab [1]{#1}%
\providecommand \enquote  [1]{``#1''}%
\providecommand \bibnamefont  [1]{#1}%
\providecommand \bibfnamefont [1]{#1}%
\providecommand \citenamefont [1]{#1}%
\providecommand \href@noop [0]{\@secondoftwo}%
\providecommand \href [0]{\begingroup \@sanitize@url \@href}%
\providecommand \@href[1]{\@@startlink{#1}\@@href}%
\providecommand \@@href[1]{\endgroup#1\@@endlink}%
\providecommand \@sanitize@url [0]{\catcode `\\12\catcode `\$12\catcode
  `\&12\catcode `\#12\catcode `\^12\catcode `\_12\catcode `\%12\relax}%
\providecommand \@@startlink[1]{}%
\providecommand \@@endlink[0]{}%
\providecommand \url  [0]{\begingroup\@sanitize@url \@url }%
\providecommand \@url [1]{\endgroup\@href {#1}{\urlprefix }}%
\providecommand \urlprefix  [0]{URL }%
\providecommand \Eprint [0]{\href }%
\providecommand \doibase [0]{http://dx.doi.org/}%
\providecommand \selectlanguage [0]{\@gobble}%
\providecommand \bibinfo  [0]{\@secondoftwo}%
\providecommand \bibfield  [0]{\@secondoftwo}%
\providecommand \translation [1]{[#1]}%
\providecommand \BibitemOpen [0]{}%
\providecommand \bibitemStop [0]{}%
\providecommand \bibitemNoStop [0]{.\EOS\space}%
\providecommand \EOS [0]{\spacefactor3000\relax}%
\providecommand \BibitemShut  [1]{\csname bibitem#1\endcsname}%
\let\auto@bib@innerbib\@empty
\bibitem [{\citenamefont {Jaynes}(1977)}]{bicameral}%
  \BibitemOpen
  \bibfield  {author} {\bibinfo {author} {\bibfnamefont {Julian}\ \bibnamefont
  {Jaynes}},\ }\href@noop {} {\enquote {\bibinfo {title} {{T}he origin of
  consciousness in the breakdown of the bicameral mind},}\ } (\bibinfo {year}
  {1977})\BibitemShut {NoStop}%
\bibitem [{\citenamefont {Merleau{-}Ponty}(1964)}]{Merleau-Ponty1964-MEREAM-2}%
  \BibitemOpen
  \bibfield  {author} {\bibinfo {author} {\bibfnamefont {Maurice}\ \bibnamefont
  {Merleau{-}Ponty}},\ }\bibfield  {title} {\enquote {\bibinfo {title} {Eye and
  {M}ind},}\ }in\ \href@noop {} {\emph {\bibinfo {booktitle} {The {P}rimacy of
  {P}erception}}}\ (\bibinfo  {publisher} {Evanston, {U}{S}{A}: {N}orthwestern
  {U}niversity {P}ress},\ \bibinfo {year} {1964})\ pp.\ \bibinfo {pages}
  {159--190}\BibitemShut {NoStop}%
\bibitem [{\citenamefont {Nagel}(1974)}]{Nagel1974-NAGWII}%
  \BibitemOpen
  \bibfield  {author} {\bibinfo {author} {\bibfnamefont {Thomas}\ \bibnamefont
  {Nagel}},\ }\bibfield  {title} {\enquote {\bibinfo {title} {{W}hat is {I}t
  {L}ike to {B}e a {B}at?}}\ }\href {\doibase 10.2307/2183914} {\bibfield
  {journal} {\bibinfo  {journal} {{P}hilosophical {R}eview}\ }\textbf {\bibinfo
  {volume} {83}},\ \bibinfo {pages} {435--50} (\bibinfo {year}
  {1974})}\BibitemShut {NoStop}%
\bibitem [{\citenamefont {Dick}(1978)}]{dick_howtobuild}%
  \BibitemOpen
  \bibfield  {author} {\bibinfo {author} {\bibfnamefont {Philip~K.}\
  \bibnamefont {Dick}},\ }\href
  {https://web.archive.org/web/20080125030037/http://deoxy.org/pkd_how2build.htm}
  {\enquote {\bibinfo {title} {How to {B}uild a {U}niverse that {D}oesn’t
  {F}all {A}part in {T}wo {D}ays.}}\ } (\bibinfo {year} {1978}),\ \bibinfo
  {note} {a speech written in 1978. {R}etrieved online June 8,
  2023}\BibitemShut {NoStop}%
\bibitem [{\citenamefont {{E}{Z}{L}{I}{F}{E}}\ and\ \citenamefont
  {{W}almart}(2023)}]{walmart_teeth}%
  \BibitemOpen
  \bibfield  {author} {\bibinfo {author} {\bibnamefont {{E}{Z}{L}{I}{F}{E}}}\
  and\ \bibinfo {author} {\bibnamefont {{W}almart}},\ }\href
  {https://www.walmart.com/ip/JUMPING-TEETH-CHATTERING-SMILE-TEETH-Small-Wind-Up-Feet-Toy-SALE-Pro-HOT-S5O9/1729610599}
  {\enquote {\bibinfo {title} {{J}{U}{M}{P}{I}{N}{G} {T}{E}{E}{T}{H}
  {C}{H}{A}{T}{T}{E}{R}{I}{N}{G} {S}{M}{I}{L}{E} {T}{E}{E}{T}{H} {S}mall {W}ind
  {U}p {F}eet {T}oy {S}{A}{L}{E} {P}ro {H}{O}{T} {S}5o9},}\ } (\bibinfo {year}
  {2023}),\ \bibinfo {note} {{J}.{W}.{Z}. used this image as a reference in the
  {P}rocreate app (on i{P}ad) to create the micrography figure, image retrieved
  in {M}ay 2023}\BibitemShut {NoStop}%
\bibitem [{\citenamefont {Dawson}(2013)}]{d_mindbodyworld}%
  \BibitemOpen
  \bibfield  {author} {\bibinfo {author} {\bibfnamefont {Michael R.~W.}\
  \bibnamefont {Dawson}},\ }\href {\doibase
  https://doi.org/10.15215/aupress/9781927356173.01} {\emph {\bibinfo {title}
  {{M}ind, {B}ody, {W}orld: {F}oundations of {C}ognitive {S}cience}}}\
  (\bibinfo  {publisher} {{A}thabasca {U}niversity {P}ress},\ \bibinfo {year}
  {2013})\ \bibinfo {note} {{A}ccessed via the {L}ibre{T}exts
  libraries}\BibitemShut {NoStop}%
\bibitem [{\citenamefont {Kaushal}\ and\ \citenamefont
  {Mahowald}(2022)}]{kaushal2022tokens}%
  \BibitemOpen
  \bibfield  {author} {\bibinfo {author} {\bibfnamefont {Ayush}\ \bibnamefont
  {Kaushal}}\ and\ \bibinfo {author} {\bibfnamefont {Kyle}\ \bibnamefont
  {Mahowald}},\ }\href@noop {} {\enquote {\bibinfo {title} {{W}hat do tokens
  know about their characters and how do they know it?}}\ } (\bibinfo {year}
  {2022}),\ \Eprint {http://arxiv.org/abs/2206.02608} {arXiv:2206.02608
  [cs.CL]} \BibitemShut {NoStop}%
\bibitem [{\citenamefont {Li}\ \emph {et~al.}(2023)\citenamefont {Li},
  \citenamefont {Karamolegkou}, \citenamefont {Kementchedjhieva}, \citenamefont
  {Abdou}, \citenamefont {Lehmann},\ and\ \citenamefont
  {Søgaard}}]{li2023large}%
  \BibitemOpen
  \bibfield  {author} {\bibinfo {author} {\bibfnamefont {Jiaang}\ \bibnamefont
  {Li}}, \bibinfo {author} {\bibfnamefont {Antonia}\ \bibnamefont
  {Karamolegkou}}, \bibinfo {author} {\bibfnamefont {Yova}\ \bibnamefont
  {Kementchedjhieva}}, \bibinfo {author} {\bibfnamefont {Mostafa}\ \bibnamefont
  {Abdou}}, \bibinfo {author} {\bibfnamefont {Sune}\ \bibnamefont {Lehmann}}, \
  and\ \bibinfo {author} {\bibfnamefont {Anders}\ \bibnamefont {Søgaard}},\
  }\href {https://arxiv.org/abs/2306.01930} {\enquote {\bibinfo {title} {Large
  {L}anguage {M}odels {C}onverge on {B}rain-{L}ike {W}ord {R}epresentations},}\
  } (\bibinfo {year} {2023}),\ \Eprint {http://arxiv.org/abs/2306.01930}
  {arXiv:2306.01930 [cs.CL]} \BibitemShut {NoStop}%
\bibitem [{\citenamefont {Bubeck}\ \emph {et~al.}(2023)\citenamefont {Bubeck},
  \citenamefont {Chandrasekaran}, \citenamefont {Eldan}, \citenamefont
  {Gehrke}, \citenamefont {Horvitz}, \citenamefont {Kamar}, \citenamefont
  {Lee}, \citenamefont {Lee}, \citenamefont {Li}, \citenamefont {Lundberg},
  \citenamefont {Nori}, \citenamefont {Palangi}, \citenamefont {Ribeiro},\ and\
  \citenamefont {Zhang}}]{bubeck2023sparks}%
  \BibitemOpen
  \bibfield  {author} {\bibinfo {author} {\bibfnamefont {Sébastien}\
  \bibnamefont {Bubeck}}, \bibinfo {author} {\bibfnamefont {Varun}\
  \bibnamefont {Chandrasekaran}}, \bibinfo {author} {\bibfnamefont {Ronen}\
  \bibnamefont {Eldan}}, \bibinfo {author} {\bibfnamefont {Johannes}\
  \bibnamefont {Gehrke}}, \bibinfo {author} {\bibfnamefont {Eric}\ \bibnamefont
  {Horvitz}}, \bibinfo {author} {\bibfnamefont {Ece}\ \bibnamefont {Kamar}},
  \bibinfo {author} {\bibfnamefont {Peter}\ \bibnamefont {Lee}}, \bibinfo
  {author} {\bibfnamefont {Yin~Tat}\ \bibnamefont {Lee}}, \bibinfo {author}
  {\bibfnamefont {Yuanzhi}\ \bibnamefont {Li}}, \bibinfo {author}
  {\bibfnamefont {Scott}\ \bibnamefont {Lundberg}}, \bibinfo {author}
  {\bibfnamefont {Harsha}\ \bibnamefont {Nori}}, \bibinfo {author}
  {\bibfnamefont {Hamid}\ \bibnamefont {Palangi}}, \bibinfo {author}
  {\bibfnamefont {Marco~Tulio}\ \bibnamefont {Ribeiro}}, \ and\ \bibinfo
  {author} {\bibfnamefont {Yi}~\bibnamefont {Zhang}},\ }\href@noop {} {\enquote
  {\bibinfo {title} {{S}parks of {A}rtificial {G}eneral {I}ntelligence: {E}arly
  experiments with {G}{P}{T}-4},}\ } (\bibinfo {year} {2023}),\ \Eprint
  {http://arxiv.org/abs/2303.12712} {arXiv:2303.12712 [cs.CL]} \BibitemShut
  {NoStop}%
\bibitem [{\citenamefont {Yao}\ \emph {et~al.}(2023)\citenamefont {Yao},
  \citenamefont {Yu}, \citenamefont {Zhao}, \citenamefont {Shafran},
  \citenamefont {Griffiths}, \citenamefont {Cao},\ and\ \citenamefont
  {Narasimhan}}]{yao2023tree}%
  \BibitemOpen
  \bibfield  {author} {\bibinfo {author} {\bibfnamefont {Shunyu}\ \bibnamefont
  {Yao}}, \bibinfo {author} {\bibfnamefont {Dian}\ \bibnamefont {Yu}}, \bibinfo
  {author} {\bibfnamefont {Jeffrey}\ \bibnamefont {Zhao}}, \bibinfo {author}
  {\bibfnamefont {Izhak}\ \bibnamefont {Shafran}}, \bibinfo {author}
  {\bibfnamefont {Thomas~L.}\ \bibnamefont {Griffiths}}, \bibinfo {author}
  {\bibfnamefont {Yuan}\ \bibnamefont {Cao}}, \ and\ \bibinfo {author}
  {\bibfnamefont {Karthik}\ \bibnamefont {Narasimhan}},\ }\href@noop {}
  {\enquote {\bibinfo {title} {{T}ree of {T}houghts: {D}eliberate {P}roblem
  {S}olving with {L}arge {L}anguage {M}odels},}\ } (\bibinfo {year} {2023}),\
  \Eprint {http://arxiv.org/abs/2305.10601} {arXiv:2305.10601 [cs.CL]}
  \BibitemShut {NoStop}%
\bibitem [{\citenamefont {Su}\ \emph {et~al.}(2020)\citenamefont {Su},
  \citenamefont {Zhu}, \citenamefont {Cao}, \citenamefont {Li}, \citenamefont
  {Lu}, \citenamefont {Wei},\ and\ \citenamefont {Dai}}]{su2020vlbert}%
  \BibitemOpen
  \bibfield  {author} {\bibinfo {author} {\bibfnamefont {Weijie}\ \bibnamefont
  {Su}}, \bibinfo {author} {\bibfnamefont {Xizhou}\ \bibnamefont {Zhu}},
  \bibinfo {author} {\bibfnamefont {Yue}\ \bibnamefont {Cao}}, \bibinfo
  {author} {\bibfnamefont {Bin}\ \bibnamefont {Li}}, \bibinfo {author}
  {\bibfnamefont {Lewei}\ \bibnamefont {Lu}}, \bibinfo {author} {\bibfnamefont
  {Furu}\ \bibnamefont {Wei}}, \ and\ \bibinfo {author} {\bibfnamefont
  {Jifeng}\ \bibnamefont {Dai}},\ }\href@noop {} {\enquote {\bibinfo {title}
  {{V}{L}-{B}{E}{R}{T}: {P}re-training of {G}eneric {V}isual-{L}inguistic
  {R}epresentations},}\ } (\bibinfo {year} {2020}),\ \Eprint
  {http://arxiv.org/abs/1908.08530} {arXiv:1908.08530 [cs.CV]} \BibitemShut
  {NoStop}%
\bibitem [{\citenamefont {Pfeifer}\ and\ \citenamefont
  {Bongard}(2006)}]{pb_bodyshapes}%
  \BibitemOpen
  \bibfield  {author} {\bibinfo {author} {\bibfnamefont {Rolf}\ \bibnamefont
  {Pfeifer}}\ and\ \bibinfo {author} {\bibfnamefont {Josh}\ \bibnamefont
  {Bongard}},\ }\href@noop {} {\emph {\bibinfo {title} {How the {B}ody {S}hapes
  the {W}ay {W}e {T}hink: {A} {N}ew {V}iew of {I}ntelligence}}}\ (\bibinfo
  {publisher} {A {B}radford {B}ook, {M}{I}{T} {P}ress},\ \bibinfo {address}
  {London, {E}ngland and {C}ambridge, {M}assachusetts},\ \bibinfo {year}
  {2006})\BibitemShut {NoStop}%
\bibitem [{\citenamefont {Hornstein}(1998)}]{hornstein_encylopedia}%
  \BibitemOpen
  \bibfield  {author} {\bibinfo {author} {\bibfnamefont {Norbert}\ \bibnamefont
  {Hornstein}},\ }\href {https://chomsky.info/1998____/} {\enquote {\bibinfo
  {title} {Encyclopedia of {P}hilosophy: {N}oam {C}homsky},}\ } (\bibinfo
  {year} {1998}),\ \bibinfo {note} {retrieved on {J}une 8, 2023}\BibitemShut
  {NoStop}%
\bibitem [{\citenamefont {Chomsky}()}]{chomsky_archives}%
  \BibitemOpen
  \bibfield  {author} {\bibinfo {author} {\bibfnamefont {Noam}\ \bibnamefont
  {Chomsky}},\ }\href
  {https://archivesspace.mit.edu/repositories/2/resources/1305#} {\enquote
  {\bibinfo {title} {Noam {C}homsky personal archives, {M}{C}-0600},}\
  }\bibinfo {note} {{C}homsky has influenced ideas in linguistics through
  pretty much every aspect of his life; since this collection includes much of
  his work, it seems like a good proxy.}\BibitemShut {Stop}%
\bibitem [{\citenamefont {Open{A}{I}}(2023{\natexlab{a}})}]{o_chatgptmodel}%
  \BibitemOpen
  \bibfield  {author} {\bibinfo {author} {\bibnamefont {Open{A}{I}}},\ }\href
  {https://chat.openai.com} {\enquote {\bibinfo {title} {Chat{G}{P}{T} ({M}ay
  12 version) [{L}arge language model]},}\ } (\bibinfo {year}
  {2023}{\natexlab{a}})\BibitemShut {NoStop}%
\bibitem [{\citenamefont {Lakoff}\ and\ \citenamefont
  {Johnson}(2008)}]{Lakoff2008}%
  \BibitemOpen
  \bibfield  {author} {\bibinfo {author} {\bibfnamefont {George}\ \bibnamefont
  {Lakoff}}\ and\ \bibinfo {author} {\bibfnamefont {Mark}\ \bibnamefont
  {Johnson}},\ }\href@noop {} {\emph {\bibinfo {title} {Metaphors {W}e {L}ive
  {B}y}}}\ (\bibinfo  {publisher} {University of {C}hicago {P}ress},\ \bibinfo
  {year} {2008})\BibitemShut {NoStop}%
\bibitem [{\citenamefont {Searle}(1980)}]{Searle1980-SEAMBA}%
  \BibitemOpen
  \bibfield  {author} {\bibinfo {author} {\bibfnamefont {John}\ \bibnamefont
  {Searle}},\ }\bibfield  {title} {\enquote {\bibinfo {title} {Minds, {B}rains,
  and {P}rograms},}\ }\href {\doibase 10.1017/s0140525x00005756} {\bibfield
  {journal} {\bibinfo  {journal} {Behavioral and {B}rain {S}ciences}\ }\textbf
  {\bibinfo {volume} {3}},\ \bibinfo {pages} {417--57} (\bibinfo {year}
  {1980})}\BibitemShut {NoStop}%
\bibitem [{\citenamefont {Peirce}(1931)}]{Peirce1931}%
  \BibitemOpen
  \bibfield  {author} {\bibinfo {author} {\bibfnamefont {Charles~S.}\
  \bibnamefont {Peirce}},\ }\href
  {http://www.commens.org/bibliography/anthology_volume/peirce-charles-s-1931-collected-papers-charles-sanders-peirce-vol-i}
  {{\selectlanguage {English}\emph {\bibinfo {title} {{T}he {C}ollected
  {P}apers of {C}.{S}. {P}eirce, {V}ol. {I}: {T}he {P}rinciples of
  {P}hilosophy}}}},\ edited by\ \bibinfo {editor} {\bibfnamefont {Charles}\
  \bibnamefont {Hartshorne}}\ and\ \bibinfo {editor} {\bibfnamefont {Paul}\
  \bibnamefont {Weiss}}\ (\bibinfo  {publisher} {{H}arvard {U}niversity
  {P}ress},\ \bibinfo {address} {Cambridge},\ \bibinfo {year} {1931})\ \bibinfo
  {note} {{F}rom the {C}ommens {B}ibliography}\BibitemShut {NoStop}%
\bibitem [{\citenamefont
  {Open{A}{I}}(2023{\natexlab{b}})}]{chatgptmodel_may24}%
  \BibitemOpen
  \bibfield  {author} {\bibinfo {author} {\bibnamefont {Open{A}{I}}},\ }\href
  {https://chat.openai.com} {\enquote {\bibinfo {title} {Chat{G}{P}{T} ({M}ay
  24 version) [{L}arge language model]},}\ } (\bibinfo {year}
  {2023}{\natexlab{b}}),\ \bibinfo {note} {accessed from an account upgraded to
  {P}lus.}\BibitemShut {Stop}%
\bibitem [{\citenamefont {Harby}(2023)}]{translation}%
  \BibitemOpen
  \bibfield  {author} {\bibinfo {author} {\bibfnamefont {Abbie}\ \bibnamefont
  {Harby}},\ }\href
  {https://slator.com/resources/can-i-use-chatgpt-for-translation/} {\enquote
  {\bibinfo {title} {{C}an {I} {U}se {C}hat{G}{P}{T} for {T}ranslation?}}\ }
  (\bibinfo {year} {2023}),\ \bibinfo {note} {accessed on {M}ay 26,
  2023}\BibitemShut {NoStop}%
\bibitem [{\citenamefont {Open{A}{I}}(2023{\natexlab{c}})}]{o_gpt4model}%
  \BibitemOpen
  \bibfield  {author} {\bibinfo {author} {\bibnamefont {Open{A}{I}}},\ }\href
  {https://chat.openai.com} {\enquote {\bibinfo {title} {{G}{P}{T}-4 ({M}ay 24
  version) [{L}arge language model]},}\ } (\bibinfo {year}
  {2023}{\natexlab{c}}),\ \bibinfo {note} {the user's account has to be
  upgraded to a paid version in order to access {G}{P}{T}-4 through this
  interface.}\BibitemShut {Stop}%
\bibitem [{\citenamefont {{F}oundation {I}nc.}(2023{\natexlab{a}})}]{a249572}%
  \BibitemOpen
  \bibfield  {author} {\bibinfo {author} {\bibfnamefont {O{E}{I}{S}}\
  \bibnamefont {{F}oundation {I}nc.}},\ }\href {https://oeis.org/A249572}
  {\enquote {\bibinfo {title} {Entry {A}249572: {L}east positive integer whose
  decimal digits divide the plane into n+1 regions. {E}quivalently, least
  positive integer with n holes in its decimal digits.}}\ } (\bibinfo {year}
  {2023}{\natexlab{a}}),\ \bibinfo {note} {retrieved on {M}ay 24, 2023 from
  {N}.{J}.{A}.{S}loane's {O}n-{L}ine {E}ncyclopedia of {I}nteger
  {S}equences}\BibitemShut {NoStop}%
\bibitem [{\citenamefont {Popham}\ \emph {et~al.}(2021)\citenamefont {Popham},
  \citenamefont {Huth}, \citenamefont {Bilenko}, \citenamefont {Deniz},
  \citenamefont {Gao}, \citenamefont {Nunez-Elizalde},\ and\ \citenamefont
  {Gallant}}]{phb_visualandlinguisticsemanticrepresentations}%
  \BibitemOpen
  \bibfield  {author} {\bibinfo {author} {\bibfnamefont {Sara~F.}\ \bibnamefont
  {Popham}}, \bibinfo {author} {\bibfnamefont {Alexander~G.}\ \bibnamefont
  {Huth}}, \bibinfo {author} {\bibfnamefont {Natalia~Y.}\ \bibnamefont
  {Bilenko}}, \bibinfo {author} {\bibfnamefont {Fatma}\ \bibnamefont {Deniz}},
  \bibinfo {author} {\bibfnamefont {James~S.}\ \bibnamefont {Gao}}, \bibinfo
  {author} {\bibfnamefont {Anwar~O.}\ \bibnamefont {Nunez-Elizalde}}, \ and\
  \bibinfo {author} {\bibfnamefont {Jack~L.}\ \bibnamefont {Gallant}},\
  }\bibfield  {title} {\enquote {\bibinfo {title} {{V}isual and linguistic
  semantic representations are aligned at the border of human visual cortex},}\
  }\href {\doibase https://doi.org/10.1038/s41593-021-00921-6} {\bibfield
  {journal} {\bibinfo  {journal} {{N}ature {N}euroscience}\ }\textbf {\bibinfo
  {volume} {24}},\ \bibinfo {pages} {1628--1636} (\bibinfo {year}
  {2021})}\BibitemShut {NoStop}%
\bibitem [{\citenamefont {Juffs}\ and\ \citenamefont
  {Rodriguez}(2014)}]{juffsrodriguez_secondlanguage}%
  \BibitemOpen
  \bibfield  {author} {\bibinfo {author} {\bibfnamefont {Alan}\ \bibnamefont
  {Juffs}}\ and\ \bibinfo {author} {\bibfnamefont {Guillermo~A.}\ \bibnamefont
  {Rodriguez}},\ }\href@noop {} {\emph {\bibinfo {title} {Second {L}anguage
  {S}entence {P}rocessing}}}\ (\bibinfo  {publisher} {Routledge},\ \bibinfo
  {address} {New {Y}ork},\ \bibinfo {year} {2014})\BibitemShut {NoStop}%
\bibitem [{\citenamefont {Mann}\ and\ \citenamefont
  {Open{A}{I}}(2020)}]{gh_gpt3}%
  \BibitemOpen
  \bibfield  {author} {\bibinfo {author} {\bibfnamefont {Benjamin}\
  \bibnamefont {Mann}}\ and\ \bibinfo {author} {\bibnamefont {Open{A}{I}}},\
  }\href {https://github.com/openai/gpt-3/tree/master/data} {\enquote {\bibinfo
  {title} {{G}ithub repository for {G}{P}{T}-3},}\ } (\bibinfo {year} {2020}),\
  \bibinfo {note} {``{T}his repository has been archived by the owner on Sep
  19, 2020. {I}t is now read-only,'' accessed on June 4, 2023}\BibitemShut
  {NoStop}%
\bibitem [{\citenamefont {Wang}\ \emph {et~al.}(2023)\citenamefont {Wang},
  \citenamefont {Luo}, \citenamefont {Wang},\ and\ \citenamefont
  {Yan}}]{wang2023bot}%
  \BibitemOpen
  \bibfield  {author} {\bibinfo {author} {\bibfnamefont {Hong}\ \bibnamefont
  {Wang}}, \bibinfo {author} {\bibfnamefont {Xuan}\ \bibnamefont {Luo}},
  \bibinfo {author} {\bibfnamefont {Weizhi}\ \bibnamefont {Wang}}, \ and\
  \bibinfo {author} {\bibfnamefont {Xifeng}\ \bibnamefont {Yan}},\ }\href
  {https://arxiv.org/pdf/2305.06424.pdf} {\enquote {\bibinfo {title} {{B}ot or
  {H}uman? {D}etecting {C}hat{G}{P}{T} {I}mposters with {A} {S}ingle
  {Q}uestion},}\ } (\bibinfo {year} {2023}),\ \Eprint
  {http://arxiv.org/abs/2305.06424} {arXiv:2305.06424 [cs.CL]} \BibitemShut
  {NoStop}%
\bibitem [{\citenamefont {Shane}(2023)}]{ticktacktoe}%
  \BibitemOpen
  \bibfield  {author} {\bibinfo {author} {\bibfnamefont {Janelle}\ \bibnamefont
  {Shane}},\ }\href
  {https://www.aiweirdness.com/optimum-tic-tac-toe/?ref=ai-weirdness-newsletter}
  {\enquote {\bibinfo {title} {Optimum tic-tac-toe},}\ } (\bibinfo {year}
  {2023}),\ \bibinfo {note} {email newsletter received {M}ay 26}\BibitemShut
  {NoStop}%
\bibitem [{\citenamefont {Open{A}{I}}(2023{\natexlab{d}})}]{quote_gpt4}%
  \BibitemOpen
  \bibfield  {author} {\bibinfo {author} {\bibnamefont {Open{A}{I}}},\ }\href
  {https://openai.com/product/gpt-4} {\enquote {\bibinfo {title} {Website
  blurb},}\ } (\bibinfo {year} {2023}{\natexlab{d}}),\ \bibinfo {note}
  {accessed in {M}ay and {J}une of 2023, most recently on {J}une 2}\BibitemShut
  {NoStop}%
\bibitem [{\citenamefont {Adams}(2007)}]{hitchhikersguide}%
  \BibitemOpen
  \bibfield  {author} {\bibinfo {author} {\bibfnamefont {Douglas}\ \bibnamefont
  {Adams}},\ }\href@noop {} {\emph {\bibinfo {title} {The Hitchhiker's Guide to
  the Galaxy}}}\ (\bibinfo  {publisher} {Random House},\ \bibinfo {address}
  {New York, NY},\ \bibinfo {year} {2007})\BibitemShut {NoStop}%
\bibitem [{\citenamefont {Pearson}\ and\ \citenamefont
  {Kosslyn}(2015)}]{pk_heterogeneity}%
  \BibitemOpen
  \bibfield  {author} {\bibinfo {author} {\bibfnamefont {J}~\bibnamefont
  {Pearson}}\ and\ \bibinfo {author} {\bibfnamefont {SM}~\bibnamefont
  {Kosslyn}},\ }\bibfield  {title} {\enquote {\bibinfo {title} {{T}he
  heterogeneity of mental representation: {E}nding the imagery debate},}\
  }\href {\doibase 10.1073/pnas.1504933112} {\ \textbf {\bibinfo {volume}
  {112}} (\bibinfo {year} {2015}),\ 10.1073/pnas.1504933112},\ \bibinfo {note}
  {{R}etrieved on {M}ay 24, 2023. {P}{M}{I}{D}: 26175024; {P}{M}{C}{I}{D}:
  {P}{M}{C}4547292}\BibitemShut {NoStop}%
\bibitem [{\citenamefont {Teng}(1998)}]{teng_depictive}%
  \BibitemOpen
  \bibfield  {author} {\bibinfo {author} {\bibfnamefont {Norman~Yujen}\
  \bibnamefont {Teng}},\ }\bibfield  {title} {\enquote {\bibinfo {title} {{T}he
  {D}epictive {N}ature of {V}isual {M}ental {I}magery},}\ }\href
  {https://www.bu.edu/wcp/Papers/Mind/MindTeng.htm} {\  (\bibinfo {year}
  {1998})},\ \bibinfo {note} {{R}etrieved on {M}ay 24, 2023}\BibitemShut
  {NoStop}%
\bibitem [{\citenamefont {Pagel}(2017)}]{pagel_language}%
  \BibitemOpen
  \bibfield  {author} {\bibinfo {author} {\bibfnamefont {Mark}\ \bibnamefont
  {Pagel}},\ }\href {\doibase https://doi.org/10.1186/s12915-017-0405-3}
  {\enquote {\bibinfo {title} {{Q}\&{A}: {W}hat is human language, when did it
  evolve and why should we care?}}\ } (\bibinfo {year} {2017}),\ \bibinfo
  {note} {{A}rticle 64}\BibitemShut {NoStop}%
\bibitem [{\citenamefont {Antoniak}\ \emph {et~al.}(2023)\citenamefont
  {Antoniak}, \citenamefont {Field}, \citenamefont {Run}, \citenamefont
  {Walsh}, \citenamefont {Klein},\ and\ \citenamefont {Sap}}]{riveter}%
  \BibitemOpen
  \bibfield  {author} {\bibinfo {author} {\bibfnamefont {Maria}\ \bibnamefont
  {Antoniak}}, \bibinfo {author} {\bibfnamefont {Anjalie}\ \bibnamefont
  {Field}}, \bibinfo {author} {\bibfnamefont {Jimin}\ \bibnamefont {Run}},
  \bibinfo {author} {\bibfnamefont {Melanie}\ \bibnamefont {Walsh}}, \bibinfo
  {author} {\bibfnamefont {Lauren~F.}\ \bibnamefont {Klein}}, \ and\ \bibinfo
  {author} {\bibfnamefont {Maarten}\ \bibnamefont {Sap}},\ }\href
  {https://maartensap.com/pdfs/antoniak2023riveter.pdf} {\enquote {\bibinfo
  {title} {{R}{I}{V}{E}{T}{E}{R}: {M}easuring {P}ower and {S}ocial {D}ynamics
  {B}etween {E}ntities},}\ } (\bibinfo {year} {2023}),\ \bibinfo {note}
  {{R}etrieved on {J}une 8, 2023}\BibitemShut {NoStop}%
\bibitem [{\citenamefont {Grice}(1989)}]{g_wayofwords}%
  \BibitemOpen
  \bibfield  {author} {\bibinfo {author} {\bibfnamefont {H.~P.}\ \bibnamefont
  {Grice}},\ }\href@noop {} {\emph {\bibinfo {title} {{S}tudies in the way of
  words}}}\ (\bibinfo  {publisher} {{H}arvard {U}niversity {P}ress},\ \bibinfo
  {address} {{C}ambridge, {M}{A}},\ \bibinfo {year} {1989})\BibitemShut
  {NoStop}%
\bibitem [{\citenamefont {Bailly}(1999)}]{bailly_believe}%
  \BibitemOpen
  \bibfield  {author} {\bibinfo {author} {\bibfnamefont {Jacques~A.}\
  \bibnamefont {Bailly}},\ }\bibfield  {title} {\enquote {\bibinfo {title}
  {What {Y}ou {S}ay, {W}hat {Y}ou {B}elieve, and {W}hat {Y}ou {M}ean},}\ }\href
  {\doibase https://doi.org/10.5840/ancientphil199919Special50} {\bibfield
  {journal} {\bibinfo  {journal} {Ancient {P}hilosophy}\ }\textbf {\bibinfo
  {volume} {19}},\ \bibinfo {pages} {65--76} (\bibinfo {year} {1999})},\
  \bibinfo {note} {special {I}ssue: {R}epresentations of {P}hilosophy {I}n the
  {C}lassical {W}orld}\BibitemShut {NoStop}%
\bibitem [{\citenamefont {{M}useum of {P}aleontology~{U}nderstanding
  {E}volution}()}]{eyeballs_evolution}%
  \BibitemOpen
  \bibfield  {author} {\bibinfo {author} {\bibfnamefont {{U}{C}}~\bibnamefont
  {{M}useum of {P}aleontology~{U}nderstanding {E}volution}},\ }\href
  {https://evolution.berkeley.edu/how-your-eye-works/} {\enquote {\bibinfo
  {title} {{H}ow your eye works},}\ }\bibinfo {note} {{A}ccessed on June 4,
  2023}\BibitemShut {NoStop}%
\bibitem [{\citenamefont {Schiebinger}\ \emph {et~al.}(2011-2021)\citenamefont
  {Schiebinger}, \citenamefont {Klinge}, \citenamefont {de~Madariaga},
  \citenamefont {Paik}, \citenamefont {Schraudner},\ and\ \citenamefont
  {(Editors)}}]{sksdmpss_genderedinnovations}%
  \BibitemOpen
  \bibfield  {author} {\bibinfo {author} {\bibfnamefont {L.}~\bibnamefont
  {Schiebinger}}, \bibinfo {author} {\bibfnamefont {I.}~\bibnamefont {Klinge}},
  \bibinfo {author} {\bibfnamefont {I.~Sánchez}\ \bibnamefont {de~Madariaga}},
  \bibinfo {author} {\bibfnamefont {H.~Y.}\ \bibnamefont {Paik}}, \bibinfo
  {author} {\bibfnamefont {M.}~\bibnamefont {Schraudner}}, \ and\ \bibinfo
  {author} {\bibfnamefont {M.~Stefanick}\ \bibnamefont {(Editors)}},\ }\href
  {https://genderedinnovations.stanford.edu/methods/language.html} {\enquote
  {\bibinfo {title} {Gendered {I}nnovations in {S}cience, {H}ealth \&
  {M}edicine, {E}ngineering and {E}nvironment},}\ } (\bibinfo {year}
  {2011-2021})\BibitemShut {NoStop}%
\bibitem [{\citenamefont {Naselaris}\ \emph {et~al.}(2015)\citenamefont
  {Naselaris}, \citenamefont {Olman}, \citenamefont {Stansbury}, \citenamefont
  {Ugurbil},\ and\ \citenamefont {Gallant}}]{nosug_voxelwiseencoding}%
  \BibitemOpen
  \bibfield  {author} {\bibinfo {author} {\bibfnamefont {Thomas}\ \bibnamefont
  {Naselaris}}, \bibinfo {author} {\bibfnamefont {Cheryl~A.}\ \bibnamefont
  {Olman}}, \bibinfo {author} {\bibfnamefont {Dustin~E.}\ \bibnamefont
  {Stansbury}}, \bibinfo {author} {\bibfnamefont {Kamil}\ \bibnamefont
  {Ugurbil}}, \ and\ \bibinfo {author} {\bibfnamefont {Jack~L.}\ \bibnamefont
  {Gallant}},\ }\bibfield  {title} {\enquote {\bibinfo {title} {{A} voxel-wise
  encoding model for early visual areas decodes mental images of remembered
  scenes},}\ }\href {\doibase 10.1016/j.neuroimage.2014.10.018.} {\bibfield
  {journal} {\bibinfo  {journal} {Neuroimage}\ }\textbf {\bibinfo {volume}
  {105}},\ \bibinfo {pages} {215--228} (\bibinfo {year} {2015})}\BibitemShut
  {NoStop}%
\bibitem [{\citenamefont {{F}oundation
  {I}nc.}(2023{\natexlab{b}})}]{oeis_classic}%
  \BibitemOpen
  \bibfield  {author} {\bibinfo {author} {\bibfnamefont {O{E}{I}{S}}\
  \bibnamefont {{F}oundation {I}nc.}},\ }\href {https://oeis.org/classic.html}
  {\enquote {\bibinfo {title} {{C}lassic {S}equences {I}n {T}he {O}n-{L}ine
  {E}ncyclopedia of {I}nteger {S}equences ({O}{E}{I}{S})},}\ } (\bibinfo {year}
  {2023}{\natexlab{b}}),\ \bibinfo {note} {retrieved on {J}une 3, 2023 from
  {N}.{J}.{A}.{S}loane's {O}n-{L}ine {E}ncyclopedia of {I}nteger
  {S}equences}\BibitemShut {NoStop}%
\bibitem [{\citenamefont {{F}oundation {I}nc.}(2023{\natexlab{c}})}]{a005150}%
  \BibitemOpen
  \bibfield  {author} {\bibinfo {author} {\bibfnamefont {O{E}{I}{S}}\
  \bibnamefont {{F}oundation {I}nc.}},\ }\href {https://oeis.org/A005150}
  {\enquote {\bibinfo {title} {Entry {A}005150: {L}ook and {S}ay sequence:
  describe the previous term! (method {A} - initial term is 1). ({F}ormerly
  {M}4780)},}\ } (\bibinfo {year} {2023}{\natexlab{c}}),\ \bibinfo {note}
  {retrieved on {J}une 3, 2023 from {N}.{J}.{A}.{S}loane's {O}n-{L}ine
  {E}ncyclopedia of {I}nteger {S}equences}\BibitemShut {NoStop}%
\bibitem [{\citenamefont {{F}oundation
  {I}nc.}(2023{\natexlab{d}})}]{julia_drafts}%
  \BibitemOpen
  \bibfield  {author} {\bibinfo {author} {\bibfnamefont {O{E}{I}{S}}\
  \bibnamefont {{F}oundation {I}nc.}},\ }\href
  {hhttps://oeis.org/draft?user=Julia\%20Zimmerman} {\enquote {\bibinfo {title}
  {Unpublished related sequence drafts},}\ } (\bibinfo {year}
  {2023}{\natexlab{d}}),\ \bibinfo {note} {unpublished drafts retrieved on
  {J}une 3, 2023 from {N}.{J}.{A}.{S}loane's {O}n-{L}ine {E}ncyclopedia of
  {I}nteger {S}equences}\BibitemShut {NoStop}%
\bibitem [{\citenamefont {Open{A}{I}}(2023{\natexlab{e}})}]{between_chatgpt}%
  \BibitemOpen
  \bibfield  {author} {\bibinfo {author} {\bibnamefont {Open{A}{I}}},\ }\href
  {https://chat.openai.com} {\enquote {\bibinfo {title} {Chat{G}{P}{T} ({M}ay
  24 version) [{L}arge language model]},}\ } (\bibinfo {year}
  {2023}{\natexlab{e}}),\ \bibinfo {note} {"what does it mean that 3 is between
  2 and 4?" prompt}\BibitemShut {NoStop}%
\bibitem [{\citenamefont {{F}oundation {I}nc.}(2023{\natexlab{e}})}]{A000079}%
  \BibitemOpen
  \bibfield  {author} {\bibinfo {author} {\bibfnamefont {O{E}{I}{S}}\
  \bibnamefont {{F}oundation {I}nc.}},\ }\href {https://oeis.org/A000079}
  {\enquote {\bibinfo {title} {Entry {A}000079: {P}owers of 2: $a(n) = 2^n$.
  ({F}ormerly {M}1129 {N}0432)},}\ } (\bibinfo {year} {2023}{\natexlab{e}}),\
  \bibinfo {note} {retrieved on {J}une 3, 2023 from {N}.{J}.{A}.{S}loane's
  {O}n-{L}ine {E}ncyclopedia of {I}nteger {S}equences}\BibitemShut {NoStop}%
\bibitem [{\citenamefont {u/Pennwisedom}(2015)}]{georgebush_reddit}%
  \BibitemOpen
  \bibfield  {author} {\bibinfo {author} {\bibnamefont {u/Pennwisedom}},\
  }\href
  {https://www.reddit.com/r/languagelearning/comments/3ejh6m/comment/ctfty1z/?utm\_source=share\&utm\_medium=web3x\&utm\_name=web3xcss\&utm\_term=1\&utm\_content=share\_button}
  {} (\bibinfo {year} {2015}),\ \bibinfo {note} {{R}etrieved on {J}une 9, 2023
  (comment label says 8 years ago, so circa 2015).}\BibitemShut {Stop}%
\bibitem [{\citenamefont {Magritte}(1929)}]{magritte_pipe}%
  \BibitemOpen
  \bibfield  {author} {\bibinfo {author} {\bibfnamefont {Rene}\ \bibnamefont
  {Magritte}},\ }\href {https://en.wikipedia.org/wiki/The_Treachery_of_Images}
  {\enquote {\bibinfo {title} {{T}he {T}reachery of {I}mages},}\ } (\bibinfo
  {year} {1929}),\ \bibinfo {note} {{C}urrently on display at the {L}os
  {A}ngeles {C}ounty {M}useum of {A}rt ({L}{A}{C}{M}{A})}\BibitemShut {NoStop}%
\bibitem [{\citenamefont {Abbott}(1952)}]{flatland}%
  \BibitemOpen
  \bibfield  {author} {\bibinfo {author} {\bibfnamefont {Edwin~Abbott}\
  \bibnamefont {Abbott}},\ }\href@noop {} {\emph {\bibinfo {title} {{F}latland:
  a {R}omance of {M}any {D}imensions}}}\ (\bibinfo  {publisher} {{D}over
  {P}ublications},\ \bibinfo {address} {{N}ew {Y}ork},\ \bibinfo {year}
  {1952})\ \bibinfo {note} {{O}riginally published in 1884.}\BibitemShut
  {Stop}%
\bibitem [{\citenamefont {Yates}(1966)}]{yates_memory}%
  \BibitemOpen
  \bibfield  {author} {\bibinfo {author} {\bibfnamefont {Frances}\ \bibnamefont
  {Yates}},\ }\href@noop {} {\emph {\bibinfo {title} {The {A}rt of {M}emory}}}\
  (\bibinfo  {publisher} {University of Chicago Press},\ \bibinfo {year}
  {1966})\ pp.\ \bibinfo {pages} {126--127}\BibitemShut {NoStop}%
\bibitem [{\citenamefont {Clark}\ and\ \citenamefont
  {Chalmers}(1998)}]{cc_extendedmind}%
  \BibitemOpen
  \bibfield  {author} {\bibinfo {author} {\bibfnamefont {Andy}\ \bibnamefont
  {Clark}}\ and\ \bibinfo {author} {\bibfnamefont {David}\ \bibnamefont
  {Chalmers}},\ }\bibfield  {title} {\enquote {\bibinfo {title} {The {E}xtended
  {M}ind},}\ }\href {\doibase 10.1017/s0140525x00005756} {\bibfield  {journal}
  {\bibinfo  {journal} {Analysis}\ }\textbf {\bibinfo {volume} {58}},\ \bibinfo
  {pages} {7--19} (\bibinfo {year} {1998})}\BibitemShut {NoStop}%
\bibitem [{\citenamefont {Pratchett}(1987)}]{p_equalrites}%
  \BibitemOpen
  \bibfield  {author} {\bibinfo {author} {\bibfnamefont {Terry}\ \bibnamefont
  {Pratchett}},\ }\href@noop {} {\emph {\bibinfo {title} {{E}qual {R}ites}}}\
  (\bibinfo  {publisher} {{N}ew {A}merican {L}ibrary},\ \bibinfo {address}
  {{N}ew {Y}ork, {N}{Y}},\ \bibinfo {year} {1987})\ \bibinfo {note}
  {{D}iscworld {V}olume 3}\BibitemShut {NoStop}%
\bibitem [{\citenamefont {Kohler}(1929)}]{Kohler1929}%
  \BibitemOpen
  \bibfield  {author} {\bibinfo {author} {\bibfnamefont {Wolfgang}\
  \bibnamefont {Kohler}},\ }\href
  {https://archive.org/details/gestaltpsycholog0000kohl/page/n5/mode/2up}
  {\emph {\bibinfo {title} {Gestalt {P}sychology}}}\ (\bibinfo  {publisher}
  {Horace {L}iveright},\ \bibinfo {address} {New York},\ \bibinfo {year}
  {1929})\BibitemShut {NoStop}%
\end{thebibliography}%

\clearpage



\end{document}